\documentclass[letterpaper]{article} 
\usepackage{aaai25}  
\usepackage{times}  
\usepackage{helvet}  
\usepackage{courier}  
\usepackage[hyphens]{url}  
\usepackage{graphicx} 
\usepackage{booktabs}
\usepackage{array}
\usepackage{makecell}
\usepackage[table]{xcolor}
\usepackage{natbib}  
\usepackage{caption} 
\usepackage{tcolorbox}
\usepackage{longtable}
\usepackage{framed} 
\usepackage{multirow}
\usepackage{listings}

\lstdefinestyle{json}{
    backgroundcolor=\color{backcolour},
    commentstyle=\color{codegreen},
    keywordstyle=\color{magenta},
    numberstyle=\tiny\color{codegray},
    stringstyle=\color{codepurple},
    basicstyle=\footnotesize,
    breakatwhitespace=false,
    breaklines=true,
    captionpos=b,
    keepspaces=true,
    numbers=left,
    showspaces=false,
    showstringspaces=false,
    showtabs=false,
}

\usepackage{algorithmic}

\usepackage{newfloat}
\usepackage{listings}
\DeclareCaptionStyle{ruled}{labelfont=normalfont,labelsep=colon,strut=off} 
\title{Improving Natural Language Understanding for LLMs via Large-Scale Instruction Synthesis}
\author{
    Lin Yuan\equalcontrib,
    Jun Xu\equalcontrib,
    Honghao Gui\equalcontrib,
    Mengshu Sun,
    Zhiqiang Zhang,
    Lei Liang,
    Jun Zhou\thanks{Corresponding author}
}
\affiliations{

    Ant Group, Hangzhou, China\\
\{huiwai.yl, xujun.xj, zhongjin.ghh, mengshu.sms, lingyao.zzq, leywar.liang, jun.zhoujun\}@antgroup.com
}

\usepackage{bibentry}

\begin{document}
\maketitle

\begin{abstract}
High-quality, large-scale instructions are crucial for aligning large language models (LLMs), however, there is a severe shortage of instruction in the field of natural language understanding (NLU). 
Previous works on constructing NLU instructions mainly focus on information extraction (IE), neglecting tasks such as machine reading comprehension, question answering, and text classification. 
Furthermore, the lack of diversity in the data has led to a decreased generalization ability of trained LLMs in other NLU tasks and a noticeable decline in the fundamental model's general capabilities. 
To address this issue, we propose Hum, a large-scale, high-quality synthetic instruction corpus for NLU tasks, designed to enhance the NLU capabilities of LLMs. 
Specifically, Hum includes IE (either close IE or open IE), machine reading comprehension, text classification, and instruction generalist tasks, thereby enriching task diversity. 
Additionally, we introduce a human-LLMs collaborative mechanism to synthesize instructions, which enriches instruction diversity  by incorporating guidelines, preference rules, and format variants. 
We conduct extensive experiments on 5 NLU tasks and 28 general capability evaluation datasets for LLMs. Experimental results show that Hum enhances the NLU capabilities of six LLMs by an average of 3.1\%, with no significant decline observed in other general capabilities.
\end{abstract}

%

\section{Introduction}

NLU is a subset of natural language processing in artificial intelligence, encompassing key tasks such as machine reading comprehension, text classification, question answering, and information extraction. 
Recently, LLMs~\cite{llama3.1,DBLP:Baichuan2} have shown impressive performance in general chat, but their language understanding ability still has shortcomings~\cite{DBLP:ChatUIE,DBLP:CMmlu,DBLP:AdaptLLM}. Supervised instruction fine-tuning is an effective method for enhancing specific capabilities of LLMs~\cite{DBLP:GoLLIE,DBLP:WizardLM}.
Technical reports from Llama3.1~\cite{llama3.1} and Qwen2~\cite{qwen2} emphasize that high-quality instruction data is essential for effectively aligning LLMs, and producing such data is crucial for improving model performance significantly. 
Recently, there have been some works on high-quality instruction synthesis.
~\citet{DBLP:MAGPIE} generates numerous queries through various templates, allowing current LLMs to produce responses. ~\citet{DBLP:InstructPT} has developed an instruction synthesis framework that converts raw pre-training text into instructional formats, substantially improving the performance of pre-trained models. Additionally, ~\citet{DBLP:WizardLM} has introduced an end-to-end framework that utilizes LLMs to create evolving synthesis instruction datasets. However, the instructions synthesized by these methods are domain-independent. Currently, there is a significant scarcity of instruction synthesis specifically for NLU tasks.
\begin{figure}[t]
    \centering
    \includegraphics[width=8.2cm]{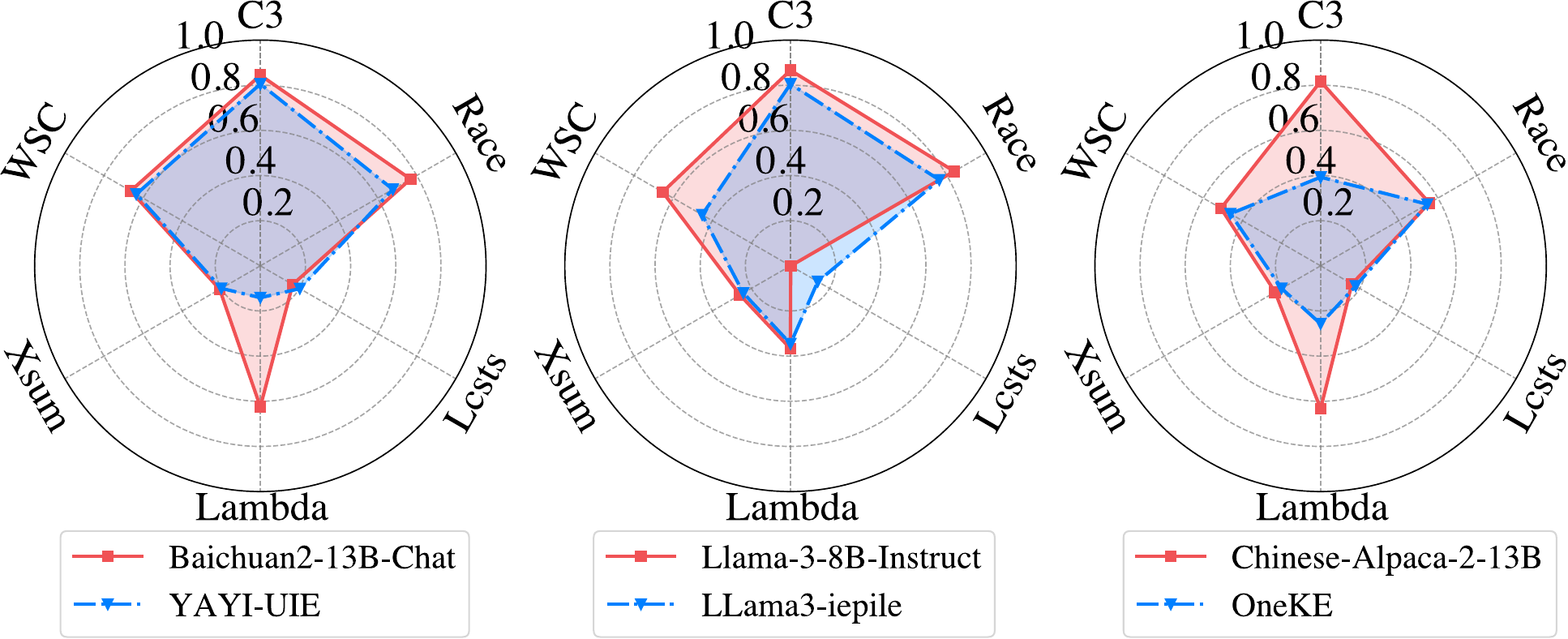}
    \caption{The existing information extraction instructions significantly reduce the performance of LLMs in NLU tasks.}
    \label{Fig.motivation}
\end{figure}

Recently, LLMs have achieved impressive performance in unified information extraction tasks by synthesizing information extraction instructions~\cite{DBLP:ChatUIE,DBLP:InstructUIE,DBLP:OneKE}. However, these synthesized instructions have several limitations.
As illustrated in Figure~\ref{Fig.motivation}, the language understanding performance of YAYI-UIE~\cite{DBLP:yayiuie}, train on the Baichuan2-13B-Chat~\cite{DBLP:Baichuan2} and YAYI datasets, and Llama3-iepile~\cite{DBLP:OneKE}, train on Llama-3-8B-Instruct~\cite{llama3.1} and IEPILE, has noticeably decreased compared to their respective LLMs, with OneKE experiencing a 14.5\% decline. An analysis reveals two main issues with the natural language understanding instructions developed in these cases: first, they primarily concentrate on information extraction tasks while overlooking machine reading comprehension, text classification, and question answering. Second, the instruction formats are too simplistic and fixed to make LLMs easily overfit these instructions. Consequently, this has significantly reduced the capabilities of LLMs trained by these instructions in the face of non-information extraction tasks or other NLU challenges.

To tackle these challenges and improve the language understanding capabilities of LLMs, we propose an effective instruction synthesis framework. First, to address the insufficient coverage of NLU tasks in prior works~\cite{DBLP:yayiuie,DBLP:OneKE,DBLP:InstructUIE,DBLP:UIE}, we construct a dataset by incorporating various tasks, including information extraction, machine reading comprehension, text classification, and instruction generalist. This increased the range of task formats from 3 to 11, significantly enhancing capabilities for non-IE tasks. Second, to overcome the limitations of previous approaches that relied on a single type of instruction, we introduced innovative instruction synthesis methods, such as guidelines synthesis, preference rules synthesis, and format variants synthesis. By utilizing a diverse array of synthesis techniques, we alleviate the issue of LLMs overfitting to a single instruction, thereby helping to maintain their proficiency in non-NLU tasks even after training on the Hum dataset. 

In summary, the main contribution of this work is as follows:
\begin{itemize}
\item We propose a framework with innovative methods for large-scale synthesis of instruction datasets. By employing guidelines synthesis, preference rules synthesis, and format variants synthesis, we address the issues of low generalization and limited instruction diversity found in NLU datasets constructed by previous works.
\item We synthesize a dataset to improve the language understanding ability of LLMs and thoroughly verified that it does not significantly impact the other general capabilities of the LLMs.
\end{itemize}

\section{Related Work}
\noindent \textbf{Generative Natural Language Understanding}
In the field of natural language understanding, mainstream tasks encompass information extraction (NER, RE, EE, OpenIE etc.), text classification (topic classification, sentiment analysis, text similarity, natural language inference, etc.), and machine reading comprehension.
For information extraction, the UIE~\cite{DBLP:UIE} framework pioneered a generation-based unified approach, effectively addressing the challenges associated with redundant models and data construction. Building on this, InstructUIE~\cite{DBLP:InstructUIE} and YAYI-UIE~\cite{DBLP:yayiuie} developed a suite of information extraction instructions, implementing an instruction-based extraction framework through fine-tuning of large language models. To further enhance generalization beyond previous extraction instructions, OneKE~\cite{DBLP:OneKE} has introduced a more comprehensive and diverse set of information extraction instructions.
In the realm of text classification, ~\citet{DBLP:zeroshotTC} and ~\citet{DBLP:GARP} have innovatively utilized different prompting methods to facilitate zeroshot text classification and natural language inference, respectively. For machine reading comprehension, ~\citet{DBLP:AdaptLLM} achieved significant performance improvements by converting extensive amounts of raw text into QA pairs before fine-tuning.

\noindent \textbf{Instruction Synthesis} 
Recent technical reports on the open-source large language models Llama 3.1~\cite{llama3.1} and Qwen2~\cite{qwen2} highlight that generating high-quality instructions is vital for training large models during both the pre-training and alignment stages. ~\citet{DBLP:Self-Instruct} proposes a framework for improving the instruction-following capabilities of pretrained language models by bootstrapping off their own generations. ~\citet{DBLP:GLAN} exclusively utilizes a pre-curated taxonomy of human knowledge and capabilities as input and generates large-scale synthetic instruction data across all disciplines. Additionally, ~\citet{DBLP:AUTOIF} transforms the validation of instruction-following data quality into code verification, requiring LLMs to generate instructions. There are also efforts to argument instructions for IE tasks. By annotation guidelines, GoLLIE~\cite{DBLP:GoLLIE} improves zero-shot information extraction, while ADELIE~\cite{DBLP:ADELIE} constructs a high-quality alignment corpus for IE instructions.

\section{Methodology}

\begin{figure*}[htbp]
    \centering
    \includegraphics[width=1\linewidth]{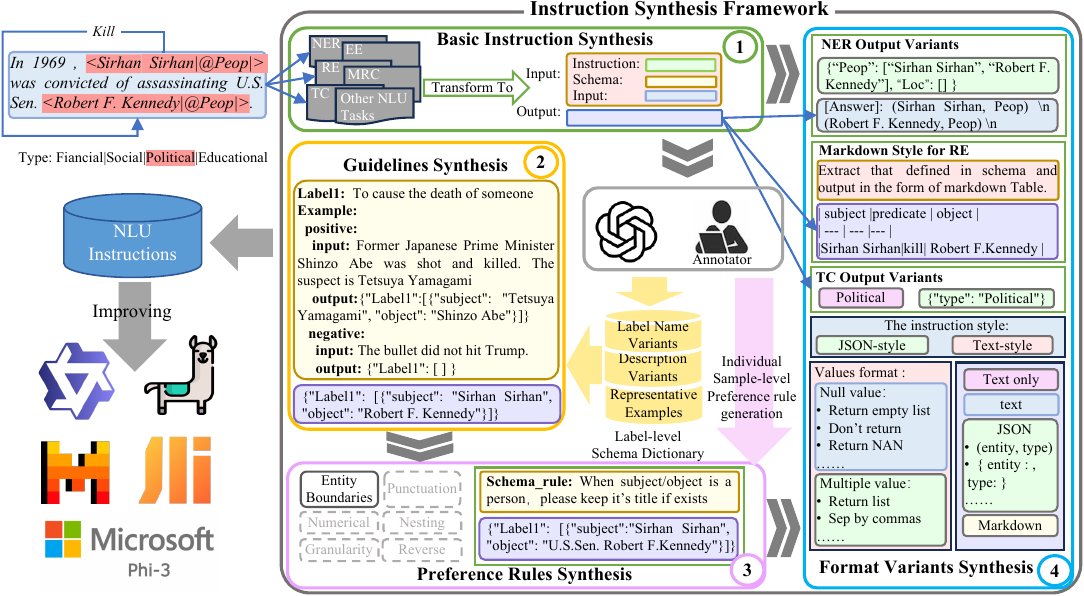}
    \caption{Overview of the natural language understanding instruction synthesis framework. The framework consists of two parts: basic instruction synthesis and compound instruction synthesis. The compound instruction synthesis mainly comprises three strategies: guidelines synthesis, preference rules synthesis, and format variants synthesis.}
    \label{Fig.framework}
\end{figure*}

The architecture of our instruction synthesis framework is illustrated in Figure~\ref{Fig.framework}, which mainly consists of two parts. First, the basic instruction synthesis, which employs the structured instruction style with the field of ``instruction", ``schema" and ``input"  from existing information extraction instructions~\cite{DBLP:UIE,DBLP:OneKE,DBLP:yayiuie,DBLP:ChatUIE} and extends to other NLU tasks, such as open information extraction, machine reading comprehension, and text classification. Second, the compound instruction synthesis, which diversifies the data from the basic instruction synthesis. The main strategies for this diversification include guidelines synthesis, preference rules synthesis, and format variants synthesis.

\subsection{Guidelines Synthesis}

Most of the previous methods~\cite{DBLP:ChatUIE,DBLP:UIE,DBLP:InstructUIE,DBLP:yayiuie} focus solely on zero-shot learning for information extraction. The instructions they design are very rigid, leading to a significant loss of in-context learning ability in large language models trained on these instructions. To address these issues, a guidelines synthesis strategy is developed. We transform the basic instructions with multiple perspectives to synthesize instructions with guidelines, which effectively prevent LLMs from overfitting and improve their language understanding capabilities. The main perspectives are as follows: 
\begin{itemize}
\item \textbf{Description:} Add semantic explanations or typical values to a schema. For example, the schema ``date'' can refer to a certain date as ``2024-08-15'' or it can also refer to a particular month or day of the week, such as ``Augest'', ``Thursday''.
\item \textbf{Example:} Providing the representative positive and negative examples in domain-specific tasks to help LLMs better follow and understand user instructions, thereby alleviating the issue of loss in-context learning capacity.
\item \textbf{Format:} Construct various structures and formats. The structures can be hierarchical or flat. The formats include JSON, text, markdown, code, etc. By specifying the output formats of an instruction and transforming the same sample into multiple corresponding structures and formats, we further reduce the overfitting of LLMs on monotonous format of instructions.
\end{itemize}

\noindent{\textbf{Guideline Paraphrasing}} To improve the generalization ability of the instructions based on the guidelines mentioned above, we make additional confusion, introduce variations, and modify the guidelines. The specific strategies are as follows:
\begin{itemize}
\item \textbf{Label Name Variants} : Utilizing synonyms to enhance the diversity of label name variants, for instance, the entity type ``Position'' may be substituted with terms such as ``Title'', ``Job'', and ``Occupation''.
\item \textbf{Label Name Masking:} Replacing a portion of the label names within the schema with placeholders in a randomized manner encourages the model to concentrate on and comprehend the schema guidelines more effectively.
\item \textbf{Description Variants}: Request LLM to generate explanations for a specific schema in a particular semantics using various expressions.
\item \textbf{Representative Examples}: For each schema, we generate five positive examples, five negative examples and various representative candidates, together with other guidelines (such as descriptions and name variants) to create a comprehensive scheme dictionary. Consequently, when synthesizing an instruction, the guidelines of a certain schema, including examples, can be randomly sampled from this dictionary.
\end{itemize}

\subsection{Preference Rules Synthesis}
\begin{figure}
    \begin{tcolorbox}[size=small, fontupper=\small, fontlower=\small]
        \textbf{Input:}
        \\ \textbf{Instruction:} You are an expert of NER. In order to improve the model's compliance with instructions, please give the description of the given schema, and generate the annotation rule, follow by which the original output meets. Then read the modification strategy and 2 example given as follows, determine whether it is possible to make slight adjustments to the origin rule to generate a new rule, and return annotated result that complies with the new rule.
        \\ \textbf{Modification Strategy:} \textit{\#Numerical (detailed content omitted) }
        \\ \textbf{Examples:} \textit{\#example (content omitted)}
        \\ \textbf{Text:} Mr. John Smith, independent director, bachelor’s degree, bachelor from Harvard, Ph.D. from MIT, senior engineer with professorship.
        \\ \textbf{Schema:} degree
        \\ \textbf{Label:} [``bachelor’s degree", ``bachelor", ``Ph.D."]
        \tcblower
        \textbf{Output:}
        \\ \textbf{Schema Description:} The name of educational qualifications and degrees.
         \\ \textbf{Original Rule:} Extract all educational qualifications and degrees of the individual.
         \\ \textbf{ New Rule:} Extract the highest educational qualification of the individual. If multiple degrees exist, annotate only the highest degree.
         \\ \textbf{ New Label:}  [``Ph.D."]
         
    \end{tcolorbox}
 \centering
 \caption{Prompt template for preference rule annotation.}
 \label{fig:Preference_Prompt}
\end{figure} 
While the synthesis of guidelines can enhance the diversity of instructions, the underlying semantics of these instructions remain almost unchanged, leading to minimal variation in model outputs. To address this limitation and synthesize samples with distinct semantics, we developed a strategy named ``preference rules synthesis'' as depicted in Figure~\ref{fig:Preference_Prompt}. This approach leverages existing guidelines to implement a modification strategy that utilizes GPT-4 for generating a novel labeling rule, subsequently producing entirely new outputs based on this rule. In contrast to the direct utilization of a rule library for invoking GPT-4 to create labeled samples, this methodology yields labeled samples with greater semantic diversity, effectively mitigating the risk of over-fitting in LLMs and improving their capability to understand fine-grained task requirements. The proposed modification strategy is outlined as follows:

\begin{itemize}
\item \textbf{Entity Boundaries:} Handling of modifying prefixes and suffixes, such as \textit{President of the United States Biden} or simply \textit{Biden}. 
\item \textbf{Numerical:} Including quantities, chronological order, logical sequence, etc. For example, extracting only the most recent position, the highest degree, or the first two companies.
\item \textbf{Granularity:} Different datasets may have varying definitions for the same schema, such as organizational entities being limited to companies only.
\item \textbf{Punctuation:} Addressing punctuation, book titles, numerical units, and so on. For example, \textit{500 \$} vs.  \textit{500}; \textit{``War and Peace"} (with quotation marks) vs. \textit{War and Peace} (without quotation marks) .
\item \textbf{Nesting:} Resolving issues of nested entities. For instance, whether \textit{Beijing} within \textit{Beijing Sport University} should be considered a geographical entity.
\item \textbf{Reverse:} The position of subject and object of a relation can be exchanged according to the semantics definition of relation label. Such as [\textit{James Cameron}, \textit{direct} (is the director of), \textit{Avatar}] VS. [\textit{Avatar}, \textit{direct} (directed by), \textit{James Cameron}].
\end{itemize}

\subsection{Format Variants Synthesis}

The previous instruction structure~\cite{DBLP:CodeIE,DBLP:OneKE,DBLP:ChatUIE} is constrained to a single output format, such as JSON, code, or plain text. However, numerous NLU tasks do not adhere to a singular representational style. For example, tasks such as machine reading comprehension and text classification do not align well with the JSON format, as they often struggle to define appropriate ``keys'' within the JSON structure. This restriction to a sole JSON instruction format poses considerable limitations on the language understanding capabilities of LLMs. To address this challenge, as show in Figure~\ref{Fig.framework}, we extend the output formats for identical samples to include JSON, text, markdown, and other styles. Additionally, we integrate prompts for various output styles within the input instructions, thereby transforming a singular sample into multiple representations. To mitigate the risk of LLMs becoming overfitted to a specific style, we generate multiple outputs for each output style. For instance, in the NER task, we define different candidates for producing empty results, such as ``'', ``NAN'', and []. By varying the format specification in the input instructions and selecting a diverse array of candidate outputs, we significantly enhance the variety of sample formats, ultimately alleviating the overfitting challenges faced by LLMs.



\subsection{Instruction Statistics}

\begin{figure}[htbp]
    \centering
    \includegraphics[width=1\linewidth]{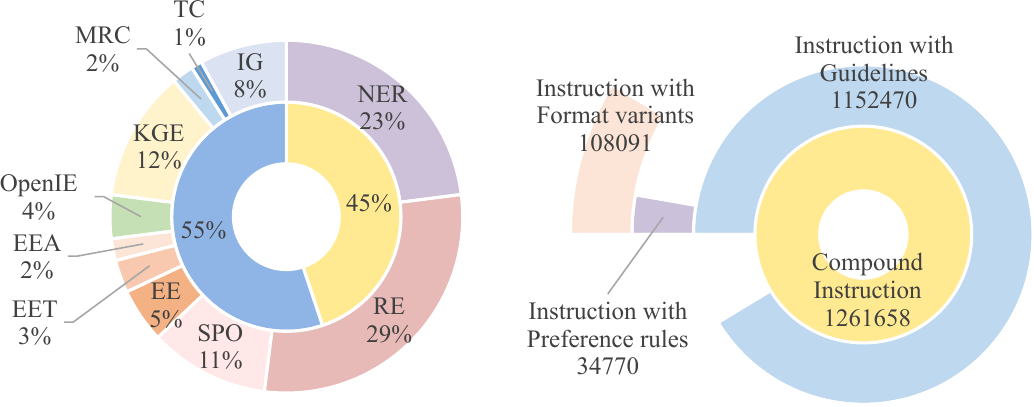}
    \caption{
    The source of the Hum dataset and the distribution of synthesis instructions.
    }
    \label{Fig.dataset}
\end{figure}

Based on the framework mentioned above, a synthesized dataset of 2,812,832 instructions is generated. As illustrated in Figure ~\ref{Fig.dataset}, the entire dataset encompasses the following tasks: NER (23\%), RE (29\%), SPO (11\%), EE (5\%), EET (3\%), EEA (2\%), OpenIE (4\%), KGE (12\%), MRC (2\%), and TC (1\%),  with an additional 8\% IG (Instruction Generalist is included to prevent LLMs from losing its chat capability). All synthesis instructions are divided into two categories: basic instructions and compound instructions. Basic instructions account for 55\% of the total. Compound instructions make up 45\% and include at least one type of instruction diversity synthesis strategy (guidelines synthesis, preference rules synthesis, format variants synthesis). The total number of compound instructions is 1,261,658, in which 1,152,470 contain guidelines, 34,770 apply preference rules synthesis, and 108,091 use format variants synthesis. Due to overlaps among these strategies, the total data volume is less than the sum of the data for each individual strategy. The definitions, examples of instructions, and data source distributions for each task, along with examples of basic and compound instructions for each task, are detailed in the Appendix.

\renewcommand\arraystretch{1.2}
\begin{table*}[htbp]
\centering
\small
\setlength\aboverulesep{0pt}\setlength\belowrulesep{0pt}
\begin{tabular}{l|cc|cc|cc|cc|cc|cc}
\toprule
                  & \multicolumn{2}{c|}{CrossNER} & \multicolumn{2}{c|}{FewRel} & \multicolumn{2}{c|}{CCF Law} & \multicolumn{2}{c|}{C3} & \multicolumn{2}{c|}{IMDB} & \multicolumn{2}{c}{Avg} \\ \cline{2-13}
                  & B           & C          & B          & C         & B           & C          & B        & C       & B         & C        & B        & C        \\ \toprule
GPT-4                 & 54.75          & 42.41          & 19.43          & 30.21          & 51.12/51.22          & 52.74/57.95          & 94.60          & 95.60          & 93.40          & 93.00          & 62.67          & 57.27          \\ 
Qwen2                 & 48.11          & 12.03          & 3.82           & 24.81          & 4.63/5.64            & 30.93/35.51          & 80.80          & 88.80          & 89.80          & 89.00          & 45.53          & 49.57          \\ 
Llama2                & 29.78          & 0.07           & 0.34           & 4.56           & 0.00/0.00                  & 12.88/11.78          & 12.00          & 39.00          & 39.60          & 78.20          & 16.34          & 26.83          \\
Baichuan2             & 40.40          & 11.10          & 2.26           & 6.72           & 0.00/0.85               & 27.64/30.32          & 39.00          & 74.20          & 83.60          & 82.40          & 33.14          & 40.68          \\ 
Llama3                & 0.00           & 0.00           & 0.00           & 0.00           & 0.00/0.00                  & 10.70/13.25           & 67.20          & 76.80          & 90.20          & 90.00          & 31.48          & 35.76          \\ \toprule
\rowcolor{red!10} OneKE                 & \textbf{52.22} & \textbf{59.94} & \textbf{33.93} & 39.14          & \textbf{62.85/62.02} & \textbf{61.86/63.88} & 68.13          & 67.04          & 82.26          & \textbf{90.28} & 59.80          & 63.85          \\ 
\rowcolor{green!10} YAYI-UIE              & 50.39          & 20.75          & \textbf{36.09} & 16.96          & 12.87/59.42          & 38.08/40.96          & \textbf{35.60} & \textbf{76.20}          & 49.40          & \textbf{92.80} & 41.53          & 45.85          \\ 
\rowcolor{blue!10} LLama3-iepile         & \textbf{51.48} & 52.30          & 23.76          & 21.71          & 56.82/57.91          & \textbf{56.73/54.54} & 51.80          & 0.80           & 86.60          & 0.00           & 54.20          & 26.09          \\ \toprule
Hum$_{\rm Qwen2}$     & 50.86          & 58.14          & 26.90          & 45.07          & 64.96/61.85          & 61.96/67.68          & 90.20          & 91.86          & 89.40          & 89.60          & 64.15          & 69.41          \\ 
\rowcolor{red!10} Hum$_{\rm Llama2}$    & 50.68          & 55.86          & 32.92          & \textbf{45.62} & 66.57/52.29          & 64.62/58.05          & \textbf{73.40} & \textbf{84.20} & \textbf{89.00} & 87.97          & \textbf{61.10} & \textbf{67.00} \\ 
\rowcolor{green!10} Hum$_{\rm Baichuan2}$ & \textbf{50.57} & \textbf{56.14} & 20.96          & \textbf{31.95} & \textbf{65.01/64.42} & \textbf{55.19/56.76} & 24.00          & 73.80 & \textbf{91.40} & 90.89          & \textbf{50.33} & \textbf{61.75} \\ 
\rowcolor{blue!10} Hum$_{\rm Llama3}$    & 49.42          & \textbf{56.41} & \textbf{31.62} & \textbf{43.39} & \textbf{62.08/61.48} & 51.63/50.85          & \textbf{82.20} & \textbf{80.80} & \textbf{92.00} & \textbf{91.00} & \textbf{63.40} & \textbf{64.57} 
\\ \toprule
\end{tabular}
\caption{Zero-shot testing for tasks related to natural language understanding. The same colored background indicates that the base model is identical. Bold text indicates that the same base model performs best. The evaluation metric used is the F1 score. B denotes basic instructions in information extraction style, and the instruction format is the same as that of OneKE and Llama3-iepile. C refers to compound instructions, which may include guidelines, rules, or multiple formats. In the CCF Law dataset, x/y represent the metrics for trigger and argument, respectively.} 
\label{Tab.overall} 
\end{table*}
\section{Experimental Settings}
\subsection{Datasets}
To evaluate the effectiveness of the Hum dataset for natural language understanding, we perform zero-shot experiments on five NLU datasets: CrossNER~\cite{DBLP:crossNER} for named entity recognition, FewRel~\cite{DBLP:FewRel} for relation extraction, CCF Law for event extraction, C3~\cite{DBLP:C3} for machine reading comprehension, and IMDB~\cite{DBLP:IMDB} for text classification. The examples of the basic and compound instructions of these five tasks are detailed in the Appendix. Additionally, to determine if the Hum dataset adversely affects LLMs, we conduct zero-shot testing across seven dimensions (language understanding, tool utilization, general knowledge, professional knowledge, coding, math, and reasoning) using a total of 28 datasets. We employ the same experimental settings as in previous work.
\subsection{Comparison Methods}
We categorize the comparison methods into two groups. The first group includes train-free models, such as GPT-4 (API), Qwen2 (Qwen2-7B-Instruct), Llama2 (Chinese-Alpaca-2-13B), Baichuan2 (Baichuan2-13B-Chat), Llama3 (Meta-Llama-3-8B-Instruct), Mistral (Mistral-7B-Instruct-v0.2), and Phi3 (Phi-3-medium-4k-instruct). The second group comprises supervised fine-tuned models like YAYI-UIE (based on Baichuan2), Llama3-iepile (based on Llama3), and OneKE (based on Llama2). \textbf{YAYI-UIE} builds upon InstructUIE~\cite{DBLP:InstructUIE} to develop a cohesive and comprehensive framework for IE instructions. This framework is subsequently refined through supervised fine-tuning with the large language model Baichuan2-13B-Chat~\cite{DBLP:Baichuan2}, resulting in a unified model capable of chat interactions. Meanwhile, \textbf{Llama3-iepile} aims to enhance the generalization capabilities of YAYI-UIE by integrating a broader variety of instructions. It has achieved better generalization on multiple datasets with Meta-Llama-3-8B-Instruct~\cite{llama3.1}. Additionally, \textbf{OneKE} utilizes the constructed IEPILE dataset and conducts extensive supervised fine-tuning with Chinese-Alpaca-2-13B~\cite{DBLP:Llama2}, leading to a LLM that demonstrates improved generalization in IE tasks. All experimental results for these models are obtained through a re-evaluation based on the officially released models and using the same instructions.
\subsection{Implementation Details}
To mitigate the impact of different models utilizing various LLMs, we perform supervised fine-tuning with Hum across multiple LLMs. We consistently apply LoRA for this fine-tuning, with a LoRA rank and alpha both set at 64 and a dropout rate of 0.05. The batch size is established at 320, accompanied by a learning rate of 5e-5. The input length is configured to 1500 tokens, while the output length is capped at 500 tokens. We utilize an Adam optimizer with weight decay at a rate of 1e-4 for training. The learning rate warm-up proportion is set to 0.1, alongside a dropout rate of 0.1. Additionally, the temperature for adjusting next token probabilities is fixed at 0.2, with the topmost probable tokens summing to a probability of 0.95. The training is conducted using the LlamaFactory~\cite{DBLP:Llamafactory} framework, leveraging 32$\times$H100 GPUs, 384 CPU cores, and 3.2TB of memory.

\section{Experimental Results}
\subsection{Overall Results}
We fine-tune the Hum dataset on LLMs and conducted zero-shot evaluations across five NLU tasks. As highlighted in Table~\ref{Tab.overall}, our model achieves superior average performance compared to other models. In the instruction tests within the basic style, it improves by 1.3\%, 8.8\%, and 9.2\% over OneKE, YAYI-UIE, and Llama3-iepile, respectively. For the compound instructions, the improvements are even more significant at 3.2
Moreover, the supervised fine-tuning performs with Qwen2 significantly enhance the performance of the Hum dataset on these NLU tasks, showing notable gains in both basic and compound instruction scenarios.
Interestingly, while OneKE showcases a commendable generalization in IE tasks such as CrossNER, FewRel, and CCF Law, its effectiveness drops sharply in non-IE NLU tasks. This dip in performance appears to stem from OneKE’s tendency to overfit the specific instructions for IE tasks, resulting in substantially lower outcomes in various other evaluations compared to Llama3-iepile and YAYI-UIE.
In contrast, models fine-tune on the Hum dataset demonstrate marked advantages over train-free LLMs across NLU tasks, regardless of whether the instructions are IE-style or compound. Among these tasks, LLMs trained on the Hum dataset (Qwen2, Llama2, Llama3) consistently surpass the performance of GPT-4.
\renewcommand\arraystretch{1.2}
\begin{table}[htbp]
\centering
\small
\setlength\aboverulesep{0pt}\setlength\belowrulesep{0pt}
\begin{tabular}{l|ccc}
\toprule
Instruction      & CrossNER & FewRel \\ \hline
basic style &  50.86     & 26.90                \\ \hline
\; + example       & 58.09                & 40.53               \\ \hline
\; + description      & 53.04                & 41.75            \\  \hline
\; + example \& description (Hum)    & 58.14                & 45.07              \\ 
\toprule
\end{tabular}
\caption{Analysis of instruction forms in in-context learning.} 
\label{Tab.ablation_model} 
\end{table}

LLMs have excellent in-context learning capabilities. By providing specific examples and descriptions, we can effectively engage the model's cognitive abilities, leading to enhanced overall performance. As indicated in Tables~\ref{Tab.overall} and~\ref{Tab.ablation_model}, when instructions are presented in the basic style (B) during inference, the model achieves moderate results on CrossNER, FewRel, and various other NLU datasets. However, incorporating examples or descriptions of the content to be extracted leads to significant performance gains on OneKE, YAYI-UIE, and Hum. Furthermore, the combination of both examples and descriptions (C) enables the model to deliver even stronger results.

\subsection{Model Ablation Studies}
We construct four datasets: one excluding guidelines synthesis instructions, one omitting preference rule synthesis instructions, one lacking format variants synthesis instructions, and one that retained all instructions. 
\renewcommand\arraystretch{1.2}
\begin{table}[htbp]
\centering
\small
\setlength\aboverulesep{0pt}\setlength\belowrulesep{0pt}
\begin{tabular}{p{1.7cm}|p{1.20cm}<{\centering}p{0.75cm}<{\centering}p{1.40cm}<{\centering}p{0.75cm}<{\centering}}
\toprule
Model                            & CrossNER & FewRel  & CCF Law & C3 \\ \toprule
    Hum$_{\rm Qwen2}$            & 58.14              & 45.07             & 61.69/67.68   & 91.86            \\ \hline
    \; - Guidelines              & 60.50               & 41.20                & 59.76/64.65   & 89.65              \\ \hline
    \; - Rules       & 57.90               & 44.36              & 58.50/56.53    & 90.40               \\ \hline
    \; - Format                  & 56.83              & 37.11              & 64.47/61.12   & 91.60               \\ 
     \toprule
\end{tabular}
\caption{Ablation experiments on various instruction data.} 
\label{Tab.ablation_data} 
\end{table}
Note that, in order to maintain the same amount of instructions as Hum, for the first three datasets, we don't simply delete the corresponding compound instructions, but instead convert them to the basic instructions. We then conduct supervised fine-tuning on these datasets using Qwen2. The results of our experiments are presented in Table~\ref{Tab.ablation_data}. Notably, the removal of preference rules has the most substantial effect on the model, leading to a marked decrease in performance across all four NLU tasks. Additionally, the absence of format variants synthesis instruction cause a significant decline in the model's performance in both CrossNER and FewRel. This indicates that integrating format variants synthesis instruction can help the LLM avoid overfitting to information extraction instructions.
Simultaneously, it is evident that these strategies exhibit minimal fluctuations in the CrossNER and C3 datasets. This can primarily be attributed to the relatively straightforward nature of these two tasks, which diminishes the observable impact of our instruction synthesis strategy.


\renewcommand\arraystretch{1.1}
\begin{table}[htbp]
\centering
\small
\setlength\aboverulesep{0pt}\setlength\belowrulesep{0pt}
\begin{tabular}{p{1.8cm}|p{0.55cm}<{\centering} p{0.55cm}<{\centering} p{0.55cm}<{\centering} p{0.75cm}<{\centering} p{0.55cm}<{\centering} p{0.55cm}<{\centering}}
\toprule
                      & C3             & WSC            & XSum           & Lambda         & Lcsts          & Race           \\ \bottomrule
GPT-4                  & 95.10          & 74.00          & 20.10          & 65.50          & 12.30          & 92.35          \\ 
\rowcolor{gray!10}Qwen2                 & 92.27          & 66.35          & 18.68          & 62.39          & 13.07          & \textbf{88.37}          \\
\rowcolor{red!10}Llama2                & 81.70           & 50.96          & 23.29          & 63.26          & 15.99          & 55.64          \\ 
\rowcolor{green!10}Baichuan2             & \textbf{84.44} & 66.35          & 20.81          & 62.43          & 16.54          & 76.85          \\ 
\rowcolor{blue!10}Llama3                & \textbf{86.63} & \textbf{65.38} & 25.84          & 36.72          & 0.09           & \textbf{83.76} \\ 
\rowcolor{yellow!10}Mistral               & \textbf{67.29} & 30.77          & 21.16          & 59.98          & 0.78           & \textbf{73.46} \\ 
\rowcolor{pink!10}Phi3                  & 68.60          & \textbf{42.31} & \textbf{0.60}  & \textbf{71.74} & 3.47           & 73.18          \\ \bottomrule
OneKE                 & 39.29          & 46.15          & 19.99          & 25.69          & 17.91          & 54.59          \\ 
YAYI-UIE              & 80.55          & 63.46          & 19.95          & 14.12          & 20.20          & 67.78          \\ 
Llama3-iepile         & 80.55          & 45.19          & 24.10          & 34.56          & 13.97          & 76.14          \\ \bottomrule
\rowcolor{gray!10}Hum$_{\rm Qwen2}$     & \textbf{92.88} & \textbf{70.19} & \textbf{31.33} & \textbf{66.16} & \textbf{18.53} & 88.17          \\
\rowcolor{red!10}Hum$_{\rm Llama2}$    & \textbf{82.36} & \textbf{63.46} & \textbf{24.51} & \textbf{65.22} & \textbf{17.51} & \textbf{68.48} \\
\rowcolor{green!10}Hum$_{\rm Baichuan2}$ & 84.11          & \textbf{66.35} & \textbf{21.51} & \textbf{62.64} & \textbf{17.27} & \textbf{77.18} \\
\rowcolor{blue!10}Hum$_{\rm Llama3}$    & 83.40          & 62.50          & \textbf{26.72} & \textbf{54.07} & \textbf{18.45} & 81.16          \\
\rowcolor{yellow!10}Hum$_{\rm Mistral}$   & 47.29          & \textbf{39.42} & \textbf{21.54} & \textbf{69.09} & \textbf{17.14} & 72.42          \\ 
\rowcolor{pink!10}Hum$_{\rm Phi3}$      & \textbf{85.21} & 25.94          & 0.36           & 71.24          & \textbf{15.49} & \textbf{74.00} \\  \bottomrule
\end{tabular}
\caption{Enhancement of natural language understanding capabilities in different LLMs by Hum. The experimental results are based on the open-compass framework and tested using  the ``gen'' mode. The evaluation metrics for C3, WSC, Lambda, and Race are ACC. XSum and Lcsts are measured using ROUGE-1. Race includes Race-middle and Race-high, and their average is taken.} 
\label{Tab.llm_nlu} 
\end{table}

\subsection{Hum For Natural Language Understanding}
\renewcommand\arraystretch{1.1}
\begin{table*}[ht]
\centering
\small
\setlength\aboverulesep{0pt}\setlength\belowrulesep{0pt}
\begin{tabular}{l|cccccccc}
\toprule
                      & \makecell[c]{Language \\ Understanding} & Tools & \makecell[c]{General \\ Knowledge} & \makecell[c]{Professional \\ Knowledge} & Coding  & Math  & Reasoning & Avg   \\ \toprule
GPT-4                  & 59.89                    & 86.44  & 78.59             & 74.23                  & 68.10  & 68.60  & 76.65     & 73.21 \\ 
\rowcolor{gray!10}Qwen2                 & 56.86                    & \textbf{76.03} & \textbf{71.52}             & \textbf{77.73}                  & \textbf{62.46} & \textbf{65.03} & \textbf{67.58}     & \textbf{68.17} \\ 
\rowcolor{red!10}Llama2                & 48.47                    & \textbf{45.68} & 51.72             & \textbf{46.98}                  & \textbf{23.37} & 17.85 & \textbf{49.65}     & 40.53 \\
\rowcolor{green!10}Baichuan2             & 54.57                    & 48.25 & \textbf{60.82}             & \textbf{55.90}                   & \textbf{25.89} & 16.62 & 45.31     & \textbf{43.91} \\ 
\rowcolor{blue!10}Llama3                & 49.74                    & 56.17 & 62.85             & 55.97                  & \textbf{55.17} & \textbf{52.37} & \textbf{59.13}     & 55.91 \\
\rowcolor{yellow!10}Mistral               & 42.24                    & 42.47 & 58.29             & \textbf{48.01}                  & 25.47 & 28.22 & 47.83     & 41.79 \\ 
\rowcolor{pink!10}Phi3                  & 43.32                    & \textbf{41.05} & 55.50              & 52.09                  & \textbf{45.23} & \textbf{63.10}  & 44.21     & 49.21 \\ \toprule
OneKE                 & 33.96                    & 30.24 & 31.85             & 31.67                  & 10.16 & 1.47  & 32.74     & 24.58 \\ YAYI-UIE              & 44.34                    & 33.89 & 55.56             & 50.58                  & 23.70  & 10.02 & \textbf{50.10}      & 38.31 \\ 
Llama3-iepile         & 45.75                    & 50.13 & 56.38             & 48.79                  & 44.96 & 46.02 & 54.36     & 49.48 \\ \toprule
\rowcolor{gray!10}Hum$_{\rm Qwen2}$     & \textbf{61.21}                    & 71.51 & 71.12             & 77.46                  & 60.48 & 59.90  & 67.33     & 67.00 \\
\rowcolor{red!10}Hum$_{\rm Llama2}$    & \textbf{53.59}                    & 45.58 & \textbf{51.90}              & 46.68                  & 21.56 & \textbf{17.96} & 49.13     & \textbf{40.91} \\
\rowcolor{green!10}Hum$_{\rm Baichuan2}$ & \textbf{54.84}                    & \textbf{49.88} & 60.77             & 55.55                  & 23.57 & \textbf{16.66} & 43.73     & 43.57 \\
\rowcolor{blue!10}Hum$_{\rm Llama3}$    & \textbf{54.38}                    & \textbf{59.71} & \textbf{64.92}             & \textbf{56.93}                  & 51.55 & 50.51 & 58.38     & \textbf{56.63} \\
\rowcolor{yellow!10}Hum$_{\rm Mistral}$   & \textbf{44.48}                    & \textbf{55.49} & \textbf{60.12}             & 47.53                  & \textbf{33.76} & \textbf{30.09} & \textbf{52.28}     & \textbf{46.25} \\ 
\rowcolor{pink!10}Hum$_{\rm Phi3}$      & \textbf{45.37}                    & 38.60  & \textbf{57.59}             & \textbf{55.96}                  & 42.42 & 61.98 & \textbf{51.14}     & \textbf{50.44} \\ \toprule
\end{tabular}
\caption{Performance evaluation of Hum in multiple dimensions across different LLMs. For each dimension, the average value of different datasets is taken as the reported value. The detailed dataset for each dimension can be found in the appendix.} 
\label{Tab.llm_overall} 
\end{table*}
We fine-tune six different LLMs using Hum data and evaluate them across seven dimensions.

As illustrated in Table~\ref{Tab.llm_overall}, models trained on the Hum dataset, such as Llama2, Llama3, Mistral, and Phi3, show an improvement in average performance across multiple dimensions. However, there is a noticeable decline in average performance for Qwen2 and Baichuan2. When comparing against models like YAYI-UIE (based on Baichuan2), Llama3-iepile (based on Llama3), and OneKE (based on Llama2), our synthesized data substantially outperformed these in multiple dimensions. Notably, tasks related to language understanding show significant improvements across all LLMs, with an average increase of 3.1\%. The models, as shown in Table~\ref{Tab.llm_nlu}, improve significantly on tasks such as Lcsts, Lambada, Xsum, and WSC, which are similar to information extraction tasks as they require extracting answers from the original text. In contrast, C3 and Race are multiple-choice question-answering tasks, and the Hum dataset lacks this type of data, leading to less noticeable results. For other dimensions, results are mixed with some showing improvements and others showing declines. It is noteworthy that in evaluations across multiple dimensions, there is no comprehensive decline observed in any single dimension. This disparity is largely attributed to our synthesized data focusing solely on language understanding, coupled with secondary SFT on instruct/chat versions of the models, which affect the general capabilities of the base models. Future work will involve synthesizing a broader variety of data to address these limitations.

\subsection{Case Study}
A typical compound instruction for relation extraction is shown in Figure ~\ref{fig:few_rel}. 
\begin{figure}[htbp]
    \begin{tcolorbox}[size=small, fontupper=\small, fontlower=\small]
        \textbf{Instruction:} You are an expert in relationship extraction. Please extract relationship triples that match the schema definition from the input. Return an empty list for relationships that do not exist. Please respond in the format of a JSON string.You can refer to the example for extraction.  
          \\ \textbf{Schema:} [\{``relation'': ``located in or next to body of water'',``description'': ``Relation between location and body of water denotes geographical connectivity. Example: Port of Hull located next to River Hull .''\}],
        \\ \textbf{Examples:} \textit{\#content omitted}

        \textbf{Input:} The Raz de Sein is bounded by the La Vieille and Petite Vieille lighthouses and by the shoreline of the le de Sein.

        \tcblower
        \textbf{Output:} 

          \par \quad   \textbf{GPT-4:} \{``located in or next to body of water": [\{``subject": ``Raz de Sein",  ``object": ``shoreline of le de Sein"\}]\}
          \par \quad  \textbf{Llama2:} The answer is too long and the content is omitted.
          \par \quad   \textbf{OneKE:} \{``located in or next to body of water'': []\}
          \par \quad  \textbf{Hum$_{\rm Llama2}$:}  \{``located in or next to body of water'': [{``subject'': ``La Vieille'', ``object'': ``Raz de Sein''}]\}
        
    \end{tcolorbox}
 \centering
 \caption{The performance of different large language models on the same compound instruction.}
 \label{fig:few_rel}
\end{figure}
The LLMs are asked to extract instances of the relation ``located in or next to body of water", the description is given in schema to indicate the semantic range of the relation: the subject is a location and the object is the body of water. Two examples are provided (due to space limitations, the content of the examples is omitted) to describe the instances that should be extracted in practice. The output format can also be determined based on the output style of examples. The results of the same input instruction from GPT-4, Llama2, OneKE and Hum$_{\rm Llama2}$ are listed. The Raz de Sein is a stretch of water, the La Vieille, Petite lighthouses lighthouses and le le de Sein are locations.
Thus in the GPT-4 result, it has made a directional error of the subject and object. For OneKE, it may unable to understand the description and examples, thus it fails to extract and relation individuals from the text. The output of Llama2 is omitted since it is too long with the chain of thought, which also makes the result hard to be parsed. Thus we thought Llama2 is failed to understand the output format from the given examples. Finally for the result of Hum$_{\rm Llama2}$, it extracts one valid relation instance and out put it in the required format.
\section{Conclusion}
In this paper, we propose a novel instruction synthesis framework to create high-quality instructions aimed at enhancing the language understanding capabilities of LLMs. We find that our synthesized Hum data significantly outperforms previous methods in NLU tasks, and notably improves the language understanding abilities of LLMs while incurring minimal knowledge loss in other dimensions.
Through ablation experiments, we discover that our proposed methods (guidelines synthesis, preference rules synthesis, and format variants synthesis) significantly enhance the model's generalization ability. Our instruction synthesis method is simple to implement and can be easily adapted for instruction synthesis across various tasks.

\bibliography{aaai25}

\appendix
\section{Hum For Fine-tuning LLMs}
The experimental results are based on the OpenCompass framework and tested using the ``gen'' mode, and the results of GPT-4 is obtained using the API. The implementation of the metric for each evaluation data can also refer to OpenCompass~\cite{Opencompass}.

\subsection{Professional Knowledge}
To evaluate the professional knowledge question-answering capabilities of large language models (LLMs) trained on Hum data, we utilized several established datasets: C-Eval~\cite{DBLP:CEval}, CMMLU~\cite{DBLP:CMmlu}, and MMLU~\cite{DBLP:MMLU}. The summarized results in Table~\ref{Tab.professional_knowledge} indicate that models like Llama3 and Phi3 demonstrated improvements with our data, whereas models such as Qwen2, Llama2, Baichuan2, and Mistral experienced declines in performance.
\renewcommand\arraystretch{1.1}
\begin{table}[htbp]
\centering
\small
\setlength\aboverulesep{0pt}\setlength\belowrulesep{0pt}
\begin{tabular}{l|cccc}
\toprule
                      & C-Eval         & CMMLU          & MMLU           & Avg            \\ \toprule
GPT-4                  & 69.63          & 70.33          & 82.74          & 74.23          \\ \hline
\rowcolor{gray!10}Qwen2                 & \textbf{81.51} & \textbf{80.95} & \textbf{70.74} & \textbf{77.73} \\ 
\rowcolor{red!10}Llama2                & 43.37          & \textbf{44.49} & \textbf{53.08} & \textbf{46.98} \\ 
\rowcolor{green!10}Baichuan2             & \textbf{55.35} & \textbf{58.26} & \textbf{54.09} & \textbf{55.90} \\ 
\rowcolor{blue!10}Llama3                & 50.39          & 50.27          & 67.25          & 55.97          \\ 
\rowcolor{yellow!10}Mistral               & \textbf{42.82} & \textbf{42.07} & 59.15          & \textbf{48.01} \\
\rowcolor{pink!10}Phi3                  & \textbf{55.84} & 22.94          & \textbf{77.50} & 52.09          \\ \toprule
OneKE                 & 32.96          & 18.67          & 43.38          & 31.67          \\  
YAYI-UIE              & 51.02          & 51.49          & 49.22          & 50.58          \\ 
Llama3-iepile         & 43.03          & 44.68          & 58.67          & 48.79          \\ \toprule
\rowcolor{gray!10}Hum$_{\rm Qwen2}$     & 81.08          & 80.88          & 70.43          & 77.46          \\ 
\rowcolor{red!10}Hum$_{\rm Llama2}$    & \textbf{44.67} & 42.49          & 52.87          & 46.68          \\ 
\rowcolor{green!10}Hum$_{\rm Baichuan2}$ & 54.66          & 58.04          & 53.94          & 55.55          \\ 
\rowcolor{blue!10}Hum$_{\rm Llama3}$    & \textbf{52.86} & \textbf{50.53} & \textbf{67.40} & \textbf{56.93} \\ 
\rowcolor{yellow!10}Hum$_{\rm Mistral}$   & 41.24          & 41.89          & \textbf{59.46} & 47.53          \\ 
\rowcolor{pink!10}Hum$_{\rm Phi3}$      & 55.66          & \textbf{34.83} & 77.40          & \textbf{55.96} \\ \toprule
\end{tabular}
\caption{Performance evaluation of professional knowledge question-answering.} 
\label{Tab.professional_knowledge} 
\end{table}
We attribute these disparities mainly to the nature of the datasets, which comprise multiple-choice exam questions. This format does not align well with the characteristics of our constructed data. Moving forward, we believe that integrating similar multiple-choice question types into our training could potentially enhance the performance of these models in professional knowledge assessment tasks.

\subsection{Coding}
To evaluate the coding capabilities of LLMs trained with Hum data, we conducted assessments using various coding datasets, such as MBPP~\cite{DBLP:MBPP} and HumanEval~\cite{DBLP:HumanEval}. The experimental results, displayed in Table \ref{Tab.code}, indicate that Hum data improved performance only in the case of Mistral, while other LLMs showed decreased performance. However, Hum-trained models still outperformed Llama3-iepile and OneKE significantly.
Our analysis suggests that LLMs trained with Hum data tend to produce code with poorer formatting, likely due to the data's bias towards the JSON format. This formatting issue could stem from the structure of Hum data, which emphasizes consistency in JSON representation, potentially at the expense of more varied coding styles and best practices typically found in traditional programming datasets. 
\renewcommand\arraystretch{1.1}
\begin{table}[ht]
\centering
\small
\setlength\aboverulesep{0pt}\setlength\belowrulesep{0pt}
\begin{tabular}{l|ccc}
\toprule
                      & MBPP           & HumanEval      & Avg            \\ \toprule
GPT-4                  & 61.80          & 74.40          & 68.10          \\ 
\rowcolor{gray!10}Qwen2                 & \textbf{54.80} & 70.12          & \textbf{62.46} \\ 
\rowcolor{red!10}Llama2                & \textbf{26.00} & \textbf{20.73} & \textbf{23.37} \\ 
\rowcolor{green!10}Baichuan2             & \textbf{28.60} & 23.17          & \textbf{25.89} \\ 
\rowcolor{blue!10}Llama3                & \textbf{52.40} & \textbf{57.93} & \textbf{55.17} \\ 
\rowcolor{yellow!10}Mistral               & 17.40          & \textbf{33.54} & 25.47          \\ 
\rowcolor{pink!10}Phi3                  & \textbf{62.40} & \textbf{28.05} & \textbf{45.23} \\ \toprule
OneKE                 & 7.60           & 12.72          & 10.16          \\  
YAYI-UIE              & 22.40          & \textbf{25.00} & 23.7           \\ 
Llama3-iepile         & 44.80          & 45.12          & 44.96          \\ \toprule
\rowcolor{gray!10}Hum$_{\rm Qwen2}$     & 49.00          & \textbf{71.95} & 60.48          \\
\rowcolor{red!10}Hum$_{\rm Llama2}$    & 23.60          & 19.51          & 21.56          \\ 
\rowcolor{green!10}Hum$_{\rm Baichuan2}$ & 26.40          & 20.73          & 23.57          \\ 
\rowcolor{blue!10}Hum$_{\rm Llama3}$    & 47.60          & 55.49          & 51.55          \\ 
\rowcolor{yellow!10}Hum$_{\rm Mistral}$   & \textbf{35.80} & 31.71          & \textbf{33.76} \\ 
\rowcolor{pink!10}Hum$_{\rm Phi3}$      & 58.00          & 26.83          & 42.42          \\ \toprule
\end{tabular}
\caption{Performance evaluation of coding.} 
\label{Tab.code}
\end{table}

\subsection{Math}
In our evaluation of the mathematical capabilities of LLMs trained with the Hum dataset, we systematically analyzed performance across various mathematical calculation benchmarks, specifically MATH~\cite{DBLP:Math} and GSM8K~\cite{DBLP:GSM8K}. As summarized in Table~\ref{Tab.math}, Llama2, Baichuan2, and Mistral exhibited significant performance improvements with the introduction of the Hum dataset. Conversely, we observed a slight decline in performance for Qwen2, Llama3, and Phi3; however, these fluctuations remained within acceptable limits.
When compared to models such as YAYI-UIE, Llama3-iepile, and OneKE, our dataset showed notable enhancements in mathematical capability. This indicates that the Hum dataset effectively addresses the limitations found in earlier datasets, particularly by increasing the diversity of the training examples.
\renewcommand\arraystretch{1.1}
\begin{table}[ht]
\centering
\small
\setlength\aboverulesep{0pt}\setlength\belowrulesep{0pt}
\begin{tabular}{l|ccc}
\toprule
                      & MATH           & GSM8K          & Avg            \\ \toprule
GPT-4                  & 45.80          & 91.40          & 68.60          \\ 
\rowcolor{gray!10}Qwen2                 & \textbf{47.42} & \textbf{82.64} & \textbf{65.03} \\ 
\rowcolor{red!10}Llama2                & 2.34           & 33.36          & 17.85          \\ \hline
\rowcolor{green!10}Baichuan2             & 7.68           & 25.55          & 16.62          \\ 
\rowcolor{blue!10}Llama3                & \textbf{26.56} & \textbf{78.17} & \textbf{52.37} \\ 
\rowcolor{yellow!10}Mistral               & 8.74           & 47.69          & 28.22          \\ 
\rowcolor{pink!10}Phi3                  & 38.10          & \textbf{88.10} & \textbf{63.10} \\ \toprule
OneKE                 & 0.06           & 2.88           & 1.47           \\ 
YAYI-UIE              & 6.62           & 13.42          & 10.02          \\ 
Llama3-iepile         & 20.08          & 71.95          & 46.02          \\ \toprule
\rowcolor{gray!10}Hum$_{\rm Qwen2}$     & 37.38          & 82.41          & 59.9           \\ 
\rowcolor{red!10}Hum$_{\rm Llama2}$    & \textbf{2.48}  & \textbf{33.43} & \textbf{17.96} \\ 
\rowcolor{green!10}Hum$_{\rm Baichuan2}$ & \textbf{8.76}  & 24.56          & \textbf{16.66} \\ 
\rowcolor{blue!10}Hum$_{\rm Llama3}$    & 24.3           & 76.72          & 50.51          \\ 
\rowcolor{yellow!10}Hum$_{\rm Mistral}$   & \textbf{9.68}  & \textbf{50.49} & \textbf{30.09} \\ 
\rowcolor{pink!10}Hum$_{\rm Phi3}$      & \textbf{38.14} & 85.82          & 61.98          \\ \toprule
\end{tabular}
\caption{Performance evaluation of mathematical calculations.} 
\label{Tab.math}
\end{table}

\subsection{Reasoning}
To evaluate the reasoning capabilities of LLMs trained with the Hum dataset, we conducted an assessment of their performance on several established reasoning benchmarks, including BBH~\cite{DBLP:BBH}, Drop~\cite{DBLP:Drop}, HellaSwag~\cite{DBLP:HellaSwag}, Ocnli~\cite{DBLP:OCNLI}, and PiQA~\cite{DBLP:PiQA}. The findings are summarized in Table ~\ref{Tab.reasoning}. Our results indicate a significant performance improvement for the Mistral and Phi3 models when utilizing the Hum dataset, while the Qwen2, Llama2, and Llama3 models exhibited a modest decline that remains within the range of normal variability. In contrast to the Llama3-iepile and OneKE models, our models demonstrated substantial enhancements, although their performance was slightly below that of YAYI-UIE. 
\renewcommand\arraystretch{1.1}
\begin{table}[ht]
\centering
\small
\setlength\aboverulesep{0pt}\setlength\belowrulesep{0pt}
\begin{tabular}{p{1.8cm}|p{0.55cm}<{\centering} p{0.55cm}<{\centering} p{0.95cm}<{\centering} p{0.55cm}<{\centering} p{0.55cm}<{\centering} p{0.55cm}<{\centering}}
\toprule
                      & BBH            & Drop           & HellaSwag      & Ocnli          & PiQA           & Avg            \\ \toprule
GPT-4                  & 88.45          & 56.00          & 91.40          & 58.20          & 89.20          & 76.65          \\ 
\rowcolor{gray!10}Qwen2                 & \textbf{64.94} & 54.82          & \textbf{78.68} & 56.71          & \textbf{82.75} & \textbf{67.58} \\ \hline
\rowcolor{red!10}Llama2                & \textbf{45.91} & \textbf{48.48} & 56.73          & 46.92          & 50.22          & \textbf{49.65} \\ 
\rowcolor{green!10}Baichuan2             & \textbf{45.73} & 27.63          & 37.13          & 48.41          & 67.63          & 45.31          \\
\rowcolor{yellow!10}\rowcolor{blue!10}Llama3                & 60.56          & \textbf{50.38} & 71.37          & 36.71          & 76.61          & \textbf{59.13} \\ 
\rowcolor{yellow!10}Mistral               & 46.50          & 7.13           & \textbf{64.12} & \textbf{48.31} & 73.07          & 47.83          \\ 
\rowcolor{pink!10}Phi3                  & \textbf{79.44} & \textbf{10.65} & \textbf{86.85} & 7.66           & 36.45          & 44.21          \\ \toprule
OneKE                 & 17.83          & 20.59          & 53.9           & 22.51          & 48.86          & 32.74          \\ 
YAYI-UIE              & 41.89          & \textbf{36.29} & \textbf{54.95} & \textbf{49.25} & \textbf{68.12} & \textbf{50.10} \\ 
Llama3-iepile         & 49.11          & 39.60          & 66.82          & 42.24          & 74.05          & 54.36          \\ \toprule
\rowcolor{gray!10}Hum$_{\rm Qwen2}$     & 62.92          & \textbf{57.52} & 76.84          & \textbf{56.85} & 82.54          & 67.33          \\ 
\rowcolor{red!10}Hum$_{\rm Llama2}$    & 45.23          & 47.56          & \textbf{57.87} & \textbf{50.88} & 44.12          & 49.13          \\ 
\rowcolor{green!10}Hum$_{\rm Baichuan2}$ & 44.43          & 21.82          & 36.65          & 48.17          & 67.57          & 43.73          \\ 
\rowcolor{blue!10}Hum$_{\rm Llama3}$    & \textbf{62.19} & 30.31          & \textbf{71.97} & \textbf{46.81} & \textbf{80.63} & 58.38          \\ 
\rowcolor{yellow!10}Hum$_{\rm Mistral}$   & \textbf{55.50} & \textbf{55.04} & 62.64          & 12.14          & \textbf{76.06} & \textbf{52.28} \\ 
\rowcolor{pink!10}Hum$_{\rm Phi3}$      & 78.18          & 3.61           & 86.00          & \textbf{50.00} & \textbf{37.92} & \textbf{51.14} \\ \toprule
\end{tabular}
\caption{Performance evaluation of reasoning.} 
\label{Tab.reasoning}
\end{table}
\renewcommand\arraystretch{1.1}
\begin{table*}[htbp]
\centering
\small
\setlength\aboverulesep{0pt}\setlength\belowrulesep{0pt}
\begin{tabular}{l|ccccccc}
\toprule
& Instruct       & Plan           & Review         & Reason         & Retrieve       & Understand     & Avg            \\ \toprule
GPT-4                  & 96.30              & 87.80             &  94.50           & 65.35             &  88.95              & 85.75              & 86.44          \\ 
\rowcolor{gray!10}Qwen2                 & \textbf{97.66} & \textbf{83.02} & 42.51          & \textbf{63.92} & \textbf{85.65} & 83.41          & \textbf{76.03} \\ 
\rowcolor{red!10}Llama2                & 57.66          & \textbf{51.75} & 46.82          & 34.58          & 41.41          & 41.85          & \textbf{45.68} \\ 
\rowcolor{green!10}Baichuan2             & 83.16          & 45.82          & 42.30          & 32.52          & 42.01          & 43.73          & 48.25          \\ 
\rowcolor{blue!10}Llama3                & 82.20          & 48.26          & 42.71          & 48.04          & 50.53          & 65.28          & 56.17          \\ 
\rowcolor{yellow!10}Mistral               & 53.50          & 62.00          & \textbf{62.01} & 30.92          & 13.48          & 32.92          & 42.47          \\
\rowcolor{pink!10}Phi3                  & 61.32          & \textbf{73.61} & \textbf{47.64} & 28.82          & 10.10          & 24.82          & \textbf{41.05} \\ \toprule
OneKE                 & 37.50          & 22.30          & 2.87           & 34.71          & 40.57          & \textbf{43.51} & 30.24          \\
YAYI-UIE              & 0.03           & 20.55          & 39.63          & \textbf{41.19} & \textbf{43.47} & \textbf{58.51} & 33.89          \\
Llama3-iepile         & \textbf{83.17} & 32.36          & \textbf{49.08} & 39.99          & 40.80          & 55.37          & 50.13          \\ \toprule
\rowcolor{gray!10}Hum$_{\rm Qwen2}$     & 86.37          & 72.85          & \textbf{42.51} & 61.46          & 82.21          & \textbf{83.70} & 71.51          \\ 
\rowcolor{red!10}Hum$_{\rm Llama2}$    & \textbf{59.67} & 40.54          & \textbf{54.21} & \textbf{34.96} & \textbf{42.00} & 42.08          & 45.58          \\
\rowcolor{green!10}Hum$_{\rm Baichuan2}$ & \textbf{86.04} & \textbf{54.02} & \textbf{42.71} & 32.00          & 40.90          & 43.60          & \textbf{49.88} \\ 
\rowcolor{blue!10}Hum$_{\rm Llama3}$    & 54.62          & \textbf{61.60} & 42.51          & \textbf{59.06} & \textbf{63.92} & \textbf{76.57} & \textbf{59.71} \\
\rowcolor{yellow!10}Hum$_{\rm Mistral}$   & \textbf{82.74} & \textbf{65.19} & 43.53          & \textbf{38.33} & \textbf{53.60} & \textbf{49.55} & \textbf{55.49} \\ 
\rowcolor{pink!10}Hum$_{\rm Phi3}$      & \textbf{65.59} & 56.96          & 36.76          & \textbf{29.01} & \textbf{16.16} & \textbf{27.11} & 38.60          \\ \toprule
\end{tabular}
\caption{Performance evaluation of tool utilization on T-Eval.} 
\label{Tab.tools} 
\end{table*}
It is important to highlight that the Hum dataset primarily targets factual data extraction, suggesting that there remains considerable room for advancement in the logical reasoning capabilities of these models.

\subsection{Tools}
To evaluate the tool utilization capabilities of LLMs trained with Hum data, we used T-Eval~\cite{Teval} as a test set. As shown in Table~\ref{Tab.tools}, models such as Baichuan2, Llama3, and Mistral demonstrated some improvement in average metrics, whereas Qwen2, Llama2, and Phi3 showed a slight decline in performance. Overall, the performance changes across these models were modest. However, when compared to models like YAYI-UIE, Llama3-iepile, and OneKE, the improvements were more significant. This enhancement can be attributed to the complex and diverse format of the data we developed. Despite this, due to the specific instructions and data format used in T-Eval, the overall performance of our Hum data in this context did not exhibit a substantial improvement.

\subsection{General Knowledge}
In our evaluation of the general knowledge question-answering capabilities of LLMs trained on Hum data, we utilized a variety of established datasets, including including ARC~\cite{DBLP:ARC}, BoolQ~\cite{DBLP:BoolQ}, GaoKao-Bench~\cite{DBLP:GaoKaoBench}, AGIEval~\cite{DBLP:AGIEval}, CommonsenseQA~\cite{DBLP:AGIEval}, NQ~\cite{DBLP:NQ}, OpenBookQA~\cite{DBLP:OpenBookQA}, and TriviaQA~\cite{DBLP:TriviaQA}. The findings, presented in Table~\ref{Tab.general_knowledge}, reveal that Llama2, Llama3, Mistral, and Phi3 demonstrated some performance improvements with our curated data. Conversely, Qwen2 and Baichuan2 showed slight declines in performance that were consistent with normal statistical fluctuations. Importantly, these changes were not statistically significant. Our analysis indicates that the Hum data appears to have a stronger emphasis on answer retrieval from existing texts rather than the generation of new content, which likely impacted the observed results.
\renewcommand\arraystretch{1.1}
\begin{table*}[htbp]
\centering
\small
\setlength\aboverulesep{0pt}\setlength\belowrulesep{0pt}
\begin{tabular}{l|cccccccccc}
\toprule
& ARC-c          & ARC-e          & BoolQ          & Gaokao         & AGIEval        & ComsenseQA  & NQ             & OpenBookQA     & TriviaQA       & Avg            \\ \toprule
GPT-4                  & 93.60          & 95.40          & 90.60          & 72.30          & 55.10          & 88.30          & 40.40          & 96.60          & 75.00          & 78.59          \\ 
\rowcolor{gray!10}Qwen2                 & \textbf{84.41} & \textbf{94.89} & 85.26          & \textbf{74.20} & \textbf{55.36} & 79.61          & 18.86          & \textbf{92.40} & 58.73          & \textbf{71.52} \\ 
\rowcolor{red!10}Llama2                & \textbf{54.24} & \textbf{71.78} & 80.73          & 25.79          & \textbf{32.82} & \textbf{52.74} & 18.75          & 68.60          & 60.04          & 51.72          \\ 
\rowcolor{green!10}Baichuan2             & \textbf{74.24} & \textbf{84.30} & 82.75          & \textbf{47.78} & 38.65          & 71.09          & 12.88          & \textbf{82.00} & 53.69          & \textbf{60.82} \\ 
\rowcolor{blue!10}Llama3                & 78.31          & 91.36          & 66.21          & \textbf{43.27} & 34.96          & 78.46          & 24.99          & 84.20          & \textbf{63.90} & 62.85          \\ 
\rowcolor{yellow!10}Mistral               & 71.86          & 81.31          & 85.69          & \textbf{29.49} & \textbf{35.07} & 71.01          & 8.12           & 82.80          & 59.26          & 58.29          \\ 
\rowcolor{pink!10}Phi3                  & 43.05          & 47.09          & \textbf{85.93} & \textbf{35.95} & \textbf{38.99} & \textbf{82.88} & 20.86          & \textbf{90.80} & 53.92          & 55.50          \\ \toprule
OneKE                 & 29.83          & 34.22          & \textbf{83.73} & 14.68          & 17.22          & 48.32          & 10.39          & 43.40          & 4.89           & 31.85          \\ 
YAYI-UIE              & 63.73          & 80.95          & 80.83          & 37.05          & 36.32          & 62.82          & 15.21          & 77.60          & 45.49          & 55.56          \\ 
Llama3-iepile         & 69.49          & 86.42          & 57.77          & 31.92          & 33.32          & 74.61          & 20.78          & 78.00          & 55.10          & 56.38          \\ \toprule
\rowcolor{gray!10}Hum$_{\rm Qwen2}$     & 82.71          & 91.89          & \textbf{86.61} & 71.29          & 52.69          & \textbf{81.08} & \textbf{23.55} & 90.80          & \textbf{59.49} & 71.12          \\ 
\rowcolor{red!10}Hum$_{\rm Llama2}$    & 50.51          & 67.37          & 82.14          & \textbf{29.76} & 30.33          & 47.99          & \textbf{24.24} & \textbf{73.80} & \textbf{60.98} & \textbf{51.90} \\ 
\rowcolor{green!10}Hum$_{\rm Baichuan2}$ & 73.90          & 84.13          & \textbf{82.78} & 44.94          & \textbf{38.82} & \textbf{71.66} & \textbf{15.60} & 81.40          & \textbf{53.74} & 60.77          \\ 
\rowcolor{blue!10}Hum$_{\rm Llama3}$    & \textbf{81.02} & \textbf{92.95} & \textbf{80.64} & 37.93          & \textbf{38.12} & \textbf{79.20} & \textbf{25.84} & \textbf{84.80} & 63.82          & \textbf{64.92} \\
\rowcolor{yellow!10}Hum$_{\rm Mistral}$   & \textbf{71.86} & \textbf{83.95} & \textbf{86.91} & 22.84          & 31.45          & \textbf{72.15} & \textbf{25.24} & \textbf{83.80} & \textbf{62.84} & \textbf{60.12} \\ 
\rowcolor{pink!10}Hum$_{\rm Phi3}$      & \textbf{50.17} & \textbf{56.97} & 84.71          & 35.34          & 33.91          & 82.06          & \textbf{26.04} & 88.80          & \textbf{60.27} & \textbf{57.59} \\ \toprule
\end{tabular}
\caption{Performance evaluation of general knowledge question-answering. BoolQ is tested using the ``ppl'' mode, the others are tested using the ``gen'' mode.} 
\label{Tab.general_knowledge} 
\end{table*}

\section{Dataset Robustness Analysis}
In this study, we employ data synthesis techniques, including guidelines, preference rules, and format variants, to generate a dataset comprising approximately 2.8 million samples. We subsequently randomly select 10K, 100K, and 1M entries from this dataset for training the Qwen2. The experimental findings are detailed in Table~\ref{Tab.close_robustness_analysis} and~\ref{Tab.open_robustness_analysis}. Our results demonstrate a positive correlation between the volume of data used for training and the NLU proficiency of the model. Furthermore, even with a modest training sample size of 10K, a notable enhancement in the model's NLU capabilities is observed.

\renewcommand\arraystretch{1.1}
\begin{table*}[htbp]
\centering
\small
\setlength\aboverulesep{0pt}\setlength\belowrulesep{0pt}
\begin{tabular}{l|cc|cc|cc|cc|cc|cc}
\toprule
                  & \multicolumn{2}{c|}{CrossNER} & \multicolumn{2}{c|}{FewRel} & \multicolumn{2}{c|}{CCF Law} & \multicolumn{2}{c|}{C3} & \multicolumn{2}{c|}{IMDB} & \multicolumn{2}{c}{Avg} \\ \cline{2-13}
                  & B           & C          & B          & C         & B           & C          & B        & C       & B         & C        & B        & C        \\\toprule
Hum(10K)  & 49.66          & 44.66          & 23.65          & 37.51          & 53.12/62.51          & 47.85/57.26          & 80.60          & 91.80          & \textbf{89.60} & 89.20          & 59.86          & 63.14          \\
Hum(100K) & 50.41          & 52.97          & 25.78          & 44.77          & 61.11/62.51          & 62.55/58.11          & 88.80          & 89.20          & \textbf{89.60} & \textbf{89.60} & 63.28          & 67.37          \\
Hum(1M)   & \textbf{51.28} & 56.61          & 26.87          & 41.76          & \textbf{67.46/64.92} & \textbf{65.11}/\textbf{68.65} & 83.40          & 86.82          & 88.80          & 89.40          & 63.30          & 68.30          \\
Hum(2.8M)      & 50.86          & \textbf{58.14} & \textbf{26.90} & \textbf{45.07} & 64.96/61.85          & 61.96/67.68          & \textbf{90.20} & \textbf{91.86} & 89.40          & \textbf{89.60} & \textbf{64.15} & \textbf{69.41}
\\ 
\toprule
\end{tabular}
\caption{Robustness testing for tasks related to natural language understanding.} 
\label{Tab.close_robustness_analysis} 
\end{table*}

\renewcommand\arraystretch{1.1}
\begin{table}[htbp]
\centering
\small
\setlength\aboverulesep{0pt}\setlength\belowrulesep{0pt}
\begin{tabular}{p{1.35cm}|p{0.5cm}<{\centering} p{0.5cm}<{\centering} p{0.5cm}<{\centering} p{0.65cm}<{\centering} p{0.5cm}<{\centering} p{0.5cm}<{\centering} p{0.5cm}<{\centering}}
\toprule
                      & C3             & WSC            & XSum           & Lambda         & Lcsts          & Race      &Avg     \\ \bottomrule
Hum(10K)   & 91.89          & \textbf{71.15} & \textbf{32.04} & 57.23          & \textbf{18.72} & 86.06          & 59.52          \\
Hum(100K) & 92.38          & 66.35          & 30.76          & 63.96          & 18.64          & 88.09          & 60.03          \\
Hum(1M) & 91.51          & 69.31          & 30.63          & 65.85          & 17.72          & 87.60          & 60.44          \\
Hum(2.8M)  & \textbf{92.88} & 70.19          & 31.33          & \textbf{66.16} & 18.53          & \textbf{88.17} & \textbf{61.21} \\
\bottomrule
\end{tabular}
\caption{Robustness testing for general natural language understanding.} 
\label{Tab.open_robustness_analysis} 
\end{table}

\section{Instruction Ablation Analysis}
We design basic instructions and compound instructions to enhance the diversity of instructions, and the statistical analysis of the instructions is shown in Figure~\ref{Fig.dataset}. We conduct an ablation analysis of the instructions on Qwen2, and the experimental results are presented in Table~\ref{Tab.close_robustness_analysis} and~\ref{Tab.open_instruction_analysis}. As shown in the tables, the performance of compound instructions is superior to that of basic instructions, and mixing basic and compound instructions yields even better overall performance. Hum enables LLM to learn to understand prompts through diverse forms of instructions, rather than merely memorizing them.

\renewcommand\arraystretch{1.1}
\begin{table*}[htbp]
\centering
\small
\setlength\aboverulesep{0pt}\setlength\belowrulesep{0pt}
\begin{tabular}{l|cc|cc|cc|cc|cc|cc}
\toprule
                  & \multicolumn{2}{c|}{CrossNER} & \multicolumn{2}{c|}{FewRel} & \multicolumn{2}{c|}{CCF Law} & \multicolumn{2}{c|}{C3} & \multicolumn{2}{c|}{IMDB} & \multicolumn{2}{c}{Avg} \\ \cline{2-13}
                  & B           & C          & B          & C         & B           & C          & B        & C       & B         & C        & B        & C        \\\toprule
Hum-B    & 50.89          & \textbf{59.25} & \textbf{27.16} & 38.82          & 64.15/\textbf{68.45} & 57.72/\textbf{68.18}          & 88.00          & 76.40          & \textbf{91.00} & \textbf{90.00} & \textbf{64.67} & 65.48          \\
Hum-C & \textbf{53.51} & 58.03          & 26.85          & 43.80          & 63.34/55.07          & \textbf{67.09}/66.55 & 86.20          & 89.40          & 89.40          & 88.60          & 63.03          & 69.33          \\
Hum          & 50.86          & 58.14          & 26.90          & \textbf{45.07} & \textbf{64.96}/61.85 & 61.96/67.68 & \textbf{90.20} & \textbf{91.86} & 89.40          & 89.60          & 64.15          & \textbf{69.41}

\\ 
\toprule
\end{tabular}
\caption{The ablation analysis of instructions related to natural language understanding tasks.} 
\label{Tab.close_instruction_analysis} 
\end{table*}

\renewcommand\arraystretch{1.1}
\begin{table}[htbp]
\centering
\small
\setlength\aboverulesep{0pt}\setlength\belowrulesep{0pt}
\begin{tabular}{p{1.35cm}|p{0.5cm}<{\centering} p{0.5cm}<{\centering} p{0.5cm}<{\centering} p{0.65cm}<{\centering} p{0.5cm}<{\centering} p{0.5cm}<{\centering} p{0.5cm}<{\centering}}
\toprule
                      & C3             & WSC            & XSum           & Lambda         & Lcsts          & Race      &Avg     \\ \bottomrule
Hum-B    & 90.03          & 56.73          & 30.86          & 63.87          & 18.21          & 87.87          & 57.93          \\
Hum-C & 92.27          & 66.35          & 29.20          & \textbf{71.16} & \textbf{19.22} & 88.02          & 61.04          \\
Hum       & \textbf{92.88} & \textbf{70.19} & \textbf{31.33} & 66.16          & 18.53          & \textbf{88.17} & \textbf{61.21}\\
\bottomrule
\end{tabular}
\caption{The ablation analysis of instructions related to general natural language understanding tasks.} 
\label{Tab.open_instruction_analysis} 
\end{table}

\section{Dataset and Instruction Explanation}
All the dataset used for instruction synthesis in this work and the count of the instruction for each task is listed in Table~\ref{Tab.hum}.
\renewcommand\arraystretch{1.1}
\begin{table*}[htbp]
\centering
\small
\setlength\aboverulesep{0pt}\setlength\belowrulesep{0pt}
\begin{tabular}{l|ll|l|ll}
\toprule
Task                  & Source                                & Count  & Task                   & Source                                 & count  \\   \hline
\multirow{24}{*}{NER}   & ACE2005~\cite{ACE05}  & 13043  & \multirow{11}{*}{RE}  &ADE~\cite{DBLP:ADE}      & 3457   \\
                 & AnatEM ~\cite{DBLP:AnatEM}  & 5710   &  & CMeIE ~\cite{DBLP:CMeIE}  & 62968  \\
                & swindle-ner * & 18753  &    & CoNLL2004 ~\cite{DBLP:CoNLL2004}                                 & 4083   \\
                & BC2GM~\cite{DBLP:bc2gm_bc4chemd}                               & 12368  & & DuIE2.0 ~\cite{DBLP:DuIE2.0}                                 & 452719 \\
                      & BC4CHEMD~\cite{DBLP:bc2gm_bc4chemd}           & 30359  &                 & GIDS ~\cite{DBLP:GIDS}                                  & 11340  \\
                      & BC5CDR ~\cite{DBLP:bc5cdr}                    & 5007   &          
                      & KBP37 ~\cite{DBLP:kbp37}                            & 36868  \\
                      & biz-query-ner*                        & 20981  &                         & NYT-RE ~\cite{DBLP:New-York-Times-RE}               & 104920 \\
                      & CCKS2017 ~\cite{ccks2017}                                & 2321   &     
                      & NYT11 ~\cite{DBLP:NYT11}                                    & 80405  \\
                      & CCKS2018 ~\cite{ccks2018}                               & 507    &        & Product taxonomy *                     & 15872  \\
                      & CLUE ~\cite{DBLP:clue}                                & 20526  &          & SciERC ~\cite{DBLP:SciERC}                                & 10659  \\
                      & CoNLL2003 ~\cite{DBLP:CoNLL2003}                               & 14055  & & SemEval ~\cite{DBLP:semval-RE}                           & 32996  \\ \cline{4-6}
                      & E-commerce-ner   *                         & 7920   & \multirow{4}{*}{SPO}    
                      & CMeIE ~\cite{DBLP:CMeIE}                                 & 36659  \\
                      & FabNER ~\cite{DBLP:FabNER}                                & 36945  &      & CoNLL2004 ~\cite{DBLP:CoNLL2004}                                & 1355   \\
                      & Financial NER *                       & 44793  &                         & DuIE2.0 ~\cite{DBLP:DuIE2.0}                                & 255807 \\
                      & FindVehicle ~\cite{DBLP:FindVehicle}           & 64935  &                 & Product taxonomy *                     & 10701  \\ \cline{4-6}
                      & GENIA ~\cite{DBLP:GENIA_NER}                             & 17411  & \multirow{7}{*}{EE}     & ACE2005~\cite{ACE05}                                  & 13507  \\
                      & HarveyNER ~\cite{DBLP:HarveyNER}                             & 6602   &   & CASIE ~\cite{DBLP:CASIE}                                 & 13309  \\
                      & MIT Movie ~\cite{DBLP:mit-movie_mit-restaurant}                & 18278  & & DuEE1.0 ~\cite{DBLP:DuEE1.0}                                 & 57997  \\
                      & MIT Restaurant ~\cite{DBLP:mit-movie_mit-restaurant}    & 16620  &        & DuEE-fin ~\cite{DBLP:DuEE-fin}                                & 29543  \\
                      & MSRA ~\cite{DBLP:MSRA}                                    & 50536  &      & IREE  ~\cite{DBLP:IREE}                                         & 6108   \\
                      & MultiNERD ~\cite{DBLP:MultiNERD}                             & 143455 &                         & PHEE  ~\cite{DBLP:PHEE}                                      & 6378   \\
                      & NCBI  ~\cite{DBLP:NCBI}                                  & 5412   &       & Publicity news classification *        & 1265   \\ \cline{4-6}
                      & Ontonotes ~\cite{DBLP:Ontonotes}                              & 92089  
                      & \multirow{5}{*}{EEA}    
                      & ACE2005~\cite{ACE05}                                  & 5078   \\
                      & RESUME ~\cite{DBLP:resume-ner}                           & 8207   & & CASIE ~\cite{DBLP:CASIE}                                 & 6539   \\ \cline{0-2}
\multirow{4}{*}{MRC}  & civil affairs Service Guide MRC *     & 5959   &                         
                      & DuEE1.0 ~\cite{DBLP:DuEE1.0}                            & 24364  \\
                      & CMRC2018 ~\cite{DBLP:CMRC2018}                  & 40788  &                & DuEE-fin ~\cite{DBLP:DuEE-fin}                              & 11928  \\
                      & KGE error detect *                          & 10000  &                         & PHEE ~\cite{DBLP:PHEE}                                     & 3714   \\ \cline{4-6}
                      & TC error detect   *             & 10000  & \multirow{5}{*}{EET}    
                      & ACE2005 ~\cite{ACE05}                                  & 9390   \\ \cline{0-2}
\multirow{5}{*}{TC}   & encyclopedic entity classification   *     & 3000   &                         
                      & CASIE ~\cite{DBLP:CASIE}                                     & 6895   \\
                      & ChnSentiCorp ~\citeyear{DBLP:CASIE}                          & 4128   &       & DuEE1.0 ~\cite{DBLP:DuEE1.0}                                & 42737  \\
                      & Intent   Classification *             & 8544   &                         & DuEE-fin ~\cite{DBLP:DuEE-fin}                                   & 15014  \\
                      & Publicity news classification *       & 2615   &                         & PHEE ~\cite{DBLP:PHEE}                                   & 3969   \\ \cline{4-6}
                      & THUCNews ~\citep{DBLP:THUCNews}                             & 5084   & \multirow{5}{*}{OpenIE} & Civil Affairs Service Guide   OpenIE * & 1865   \\ \cline{0-2}
\multirow{2}{*}{KGE}  & Encyclopedic KGE *                          & 255509 &                         
                      & ODIE ~\cite{DBLP:odie}                                   & 14266  \\
                      & InstructIE ~\cite{DBLP:InstructIE}                       & 80000  &       & OpenIE6 ~\cite{DBLP:openie6}                                   & 9000   \\ \cline{0-2}
\multirow{2}{*}{IG} & alpaca data \&  alpaca chinese data                     & 198220 &                         & Title2event ~\cite{DBLP:title2event}                              & 38081  \\
                      & COIG-CQIA ~\cite{DBLP:COIG-CQIA}                             & 40891  &  & UniNER ~\cite{DBLP:UniNER:}                                 & 44879 
 \\                     \toprule
\end{tabular}
\caption{Instruction distribution of Hum. * indicates self-built dataset.} 
\label{Tab.hum}
\end{table*}

\subsection{Named Entity Recognition (NER)}
For the NER task, we synthesized instructions based on 20 open-source datasets and 4 self-built datasets belonging to the author’s institution. Among them, swindle-ner comes from insurance, anti-fraud, and other business-related fields, biz-query-ner comes from the annotation of user queries, E-commerce-ner covers entities from E-commerce domain, Financial NER is built by annotation of entities of the financial domain from public news and financial reports. The example of instruction of NER can refer to Table~\ref{Tab.NER}.

\subsection{Relation Extraction (RE)}
For RE task, the synthesized instructions are based on 10 open-source datasets and 1 self-built dataset: product taxonomy, the construction of this dataset uses the distant supervision method to automatically label the hypernym, hyponym, synonym, and antonym relationships of products from the Baike Encyclopedia corpus. For an example of RE instruction please refer to Table~\ref{Tab.RE}.

\subsection{Triple Extraction (SPO)}
Subject-predicate-object extraction is a new task defined in this work, which is different from RE in that the constraints of subject\_type and object\_type will be specified in the schema. Consequently, the entities in extracted triples must satisfy the type constraints. 3 open-source datasets and the Product Taxonomy are used for the instruction synthesis of the SPO task. For an example of SPO instruction please refer to Table~\ref{Tab.SPO}.

\subsection{Event Extraction (EE)}
The event extract task is to extract the trigger and arguments of an event conditioned by its schema. Referring to IEPILE, the instruction template and default output format of the Event Extraction task are defined. The instruction is synthesized from 6 open-source datasets and a self-built one: the public opinion event of listed companies annotated from the news. For an example of EE instruction please refer to Table~\ref{Tab.EE}. 

\subsection{Event Argument Extraction (EEA)}
The Event Extraction of Argument task defined in this work, is to extract arguments of a event, and the trigger is already given in its schema. Five open-source datasets are used for the instruction synthesis of EEA task. For an example of EEA instruction, please refer in Table~\ref{Tab.EEA}.

\subsection{Event Trigger Extraction (EET)}
The Event Extraction of Trigger task defined in this work, is to extract the trigger of an event. The same five open-source datasets for EEA are also used for the instruction synthesis of EET task. A example of EET instruction please refer in Table~\ref{Tab.EET}.

\subsection{Open Information Extraction (OpenIE)}
Open Information Extraction is the task of generating a structured, machine-readable representation of the information in text, usually in the form of triples or n-ary propositions. Four open source and a self-built datasets are used for the instruction synthesis of this task. Note that, the required structure/format are different in each of OpenIE task. A example in given in Tabel~\ref{Tab.OpenIE}.

\subsection{Text Classification (TC)}
Text Classification is the task of assigning a label or class to a given text. Two open-source dataset and three self-built ones are used for instruction synthesis for this task. As shown in Tabel~\ref{Tab.TC}., is the candidate labels will be given in schema. ChnSentiCorp is a Chinese sentiment analysis dataset, which aims to determine the emotional attitude of a piece of text. The Intent classification is obtained by labeling user intent of the app pages based on their names and their descriptions. Encyclopedic entity Classification is automatically annotated by ChatGPT through asking the LLM to select a label from given choices for an encyclopedic entity with its description.

\subsection{Machine Reading Comprehension (MRC)}
Machine Reading Comprehension is one of the key problems in NLU, where the task is to read and comprehend a given text, and then answer questions based on it. CMRC2018 is the only open-source dataset be used, the other 3 are self-built datasets:  Civil Affairs Service Guide MRC, given a passage about a guideline for civil affair and answer question; KGE error detect: chatGPT and qwen are used to verify the annotations of KGE tasks by LLM. By the same method, the automated evaluation and correction of entity classification of encyclopedic entities are also used to synthesize MRC instructions. An example of MRC instruction can refer Table~\ref{Tab.MRC}.

\subsection{Knowledge Graph Extraction (KGE)}
Knowledge Graph Extraction task is defined to extract entities together with their properties (either data property or data property) from text with a one-pass query. Thus, it can be used for efficient KG construction. By converting the format of instructIE~\cite{DBLP:InstructIE} data, and annotating encyclopedic corpus through distant supervision and LLM, we synthesized instructions of KGE task. An example of KGE instruction can refer Table~\ref{Tab.KGE}.

\subsection{Instruction Generalist (IG)}
The addition of general instruction generalist datasets is used to prevent the model  the model from over-fitting on NLU tasks and losing its general conversational abilities. The instruction generalist includes alpaca data and alpaca Chinese data and COIG-CQIA~\cite{DBLP:COIG-CQIA}, the instructions are in their original form without any synthesis strategies.

\onecolumn



   \begin{table}
   \begin{tabular}[c]{p{1 cm}|p{15cm}}
   \toprule
    Task & Instruction 
   \\ \hline
   B  &
  \begin{lstlisting}
{
	"instruction": {
		"instruction": "You are an expert in named entity recognition. Please extract entities that match the schema definition from the input. Return an empty list if the entity type does not exist. Please respond in the format of a JSON string.",
		"schema": ["else"],
		"input": "Together with Yann LeCun, and Yoshua Bengio, Hinton won the 2018 Turing Award for conceptual and engineering breakthroughs that have made deep neural networks a critical component of computing."
	},
	"output": {"else": ["Turing Award"]}
}
  \end{lstlisting}
  \\ \hline
  C &
  \begin{lstlisting}
{
	"instruction": {
    "instruction": "You are an expert in named entity recognition. Please extract entities that match the schema definition from the input. Return an empty list if the entity type does not exist. Please respond in the format of a JSON string.You can refer to the example for extraction.",
		"schema": [{
			"entity_type": "else",
			"description": "The 'else' type includes a wide range of entities not in specific categories like objects, events, awards, or concepts. They can be names of people, movies, papers, organizations, or algorithms. This type includes anything important in a text not in other categories."}],
		"example": [{
			"input": "More recently , fictional representations of artificially intelligent robots in films such as A.I. Artificial Intelligence and Ex Machina and the 2016 TV adaptation of Westworld have engaged audience sympathy for the robots themselves .",
			"output": {
				"else": ["A.I. Artificial Intelligence", "Ex Machina", "Westworld"]}
		}, {"input": "In 1999 , Felix Gers and his advisor Jurgen Schmidhuber and Fred Cummins introduced the forget gate ( also called keep gate ) into LSTM architecture ,",
			"output": {"else": []}
		}, {"input": "Octave helps in solving linear and nonlinear problems numerically , and for performing other numerical experiments using a that is mostly compatible with MATLAB .",
			"output": {"else": []}
		}, {"input": "Eurisko made many interesting discoveries and enjoyed significant acclaim , with his paper Heuretics : Theoretical and Study of Heuristic Rules winning the Best Paper award at the 1982 Association for the Advancement of Artificial Intelligence .",
			"output": {
				"else": ["Heuretics : Theoretical and Study of Heuristic Rules", "Best Paper award"]}
		}],
		"input": "Together with Yann LeCun, and Yoshua Bengio, Hinton won the 2018 Turing Award for conceptual and engineering breakthroughs that have made deep neural networks a critical component of computing."
	},
	"output": {"else": ["Turing Award"]}
}
  \end{lstlisting}
 \\ \hline
 \end{tabular}
  \caption{Instruction Example of CrossNER} 
  \label{Tab.CrossNER}
  \end{table}

   \begin{table}
   \begin{tabular}[c]{p{1 cm}|p{15cm}}
   \toprule
    Task & Instruction 
   \\ \hline
   B &
  \begin{lstlisting}
{
	"instruction": {
		"instruction": "You are an expert in relationship extraction. Please extract relationship triples that match the schema definition from the input. Return an empty list for relationships that do not exist. Please respond in the format of a JSON string.",
		"schema": ["religion"],
	"output": {
		"religion": [{
			"subject": "Vincent Madeley Harris",
			"object": "Catholic Church"
		}]
	}
}
  \end{lstlisting}
  \\ \hline
  C  &
  \begin{lstlisting}
{
	"instruction": {
		"instruction": "You are an expert in relationship extraction. Please extract relationship triples that match the schema definition from the input. Return an empty list for relationships that do not exist. Please respond in the format of a JSON string.You can refer to the example for extraction.",
		"schema": [{
			"relation": "religion",
			"description": "This type of relation is about the connection between a subject and their religious belief or faith. The subject can be a person, organization, historical period, or group."
		}],
		"example": [{
			"input": "Leonard fought Wilfred Benitez for the WBC Welterweight Championship on November 30 , 1979 , at Caesar 's Palace in Las Vegas , Nevada .",
			"output": {
				"religion": []
			}
		}, {
			"input": "St Patrick 's Island is so called because this is where the Irish patron saint is reputed to have landed and begun his mission to convert the country to Christianity .",
			"output": {
				"religion": [{
					"subject": "patron saint",
					"object": "Christianity"
				}]
			}
		}],
		"input": "Vincent Madeley Harris ( October 14 , 1913 - March 31 , 1988 ) was an American clergyman of the Catholic Church ( Roman Rite ) ."
	},
	"output": {
		"religion": [{
			"subject": "Vincent Madeley Harris",
			"object": "Catholic Church"
		}]
	}
}
  \end{lstlisting}
 \\ \hline
 \end{tabular}
  \caption{Instruction Example of FewRel} 
  \label{Tab.FewRel}
  \end{table}

   \begin{table}
   \begin{tabular}[c]{p{1 cm}|p{15cm}}
   \toprule
    Task & Instruction 
   \\ \hline
   B  &
 
 \begin{lstlisting}

{
		"instruction": "You are an expert specializing in event extraction. Please extract events that conform to the schema definition from the input. Return NAN for non-existent arguments, and return a list if there are multiple values for an argument. Please answer in the format of a JSON string. You can refer to the example for extraction.",
		"schema": [{
			"event_type": "parenting",
			"trigger": true,
			"description": "Parenting refers to the care and upbringing of children, including providing support and care in aspects such as life, education, and emotions.",
			"arguments": [{
				"argument": "caregiver",
				"description": "Someone who cares for, looks after, supervises, or takes care of others, including parents, guardians, caregivers, etc."
			}, {
				"argument": "child",
				"description": "A child refers to a minor human being, usually the biological or adopted child of parents."
			}]
		}],
	"output": {
		"marriage": [{
			"trigger": "marriage",
			"arguments": {
				"husband": "Ning Shan",
				"wife": "Ren Xiao",
				"time": "January 6, 2013"
			}
		}],
		"birth": [{
			"trigger": "birth",
			"arguments": {
				"child": "Ning Ri",
				"time": "October 6, 2013"
			}
		}],
	"separation": [{
			"trigger": "separately",
			"arguments": {
				"husband": "Ning Shan",
				"wife": "Ren Xiao",
				"time": "July 2015"
			}
		}]
	}
}
  \end{lstlisting}
 \\ \hline
 \end{tabular}
  \caption{Basic Instruction Example of CCF Law} 
  \label{Tab.FewRel}
  \end{table}

\begin{table}
\begin{tabular}[c]{p{1 cm}|p{15cm}}
\toprule
Task & Instruction 
\\ \hline
 C &
   \begin{lstlisting}
  {
		"instruction": "You are an expert specializing in event extraction. Please extract events that conform to the schema definition from the input. Return NAN for non-existent arguments, and return a list if there are multiple values for an argument. Please answer in the format of a JSON string. You can refer to the example for extraction.",
		"schema": [{
			"event_type": "parenting",
			"trigger": true,
			"description": "Parenting refers to the care and upbringing of children, including providing support and care in aspects such as life, education, and emotions.",
			"arguments": [{
				"argument": "caregiver",
				"description": "Someone who cares for, looks after,  etc."
			}, ...]
		}],
		"example": [{
			"input": "The plaintiff Bi Xiwu and the defendant husband ...",
			"output": {
				"parenting": [{
					"trigger": "care",
					"arguments": {
						"caregiver": "Bi Xiwu",
						"child": "Zhou Fang"
					}
				}]
			}
		}, ...],
		"input": "Ren Xiao (female) and Ning Shan (male) got married on January 6, 2013 through a matchmaker, but due to the lack of understanding before marriage and incompatible personalities after marriage, they often quarreled over trivial matters. ..."
	,
	"output": {
		"marriage": [{
			"trigger": "marriage",
			"arguments": {
				"husband": "Ning Shan",
				"wife": "Ren Xiao",
				"time": "January 6, 2013"
			}
		}],
		"birth": [{
			"trigger": "birth",
			"arguments": {
				"child": "Ning Ri",
				"time": "October 6, 2013"
			}
		}],
	"separation": [{
			"trigger": "separately",
			"arguments": {
				"husband": "Ning Shan",
				"wife": "Ren Xiao",
				"time": "July 2015"
			}
		}]
	}
}
   \end{lstlisting}
   \\ \hline 
\end{tabular}
\caption{Compound Instruction Example of CCF Law}\label{Tab.CCF}
\end{table}

   \begin{table}
   \begin{tabular}[c]{p{1 cm}|p{15cm}}
   \toprule
    Task & Instruction 
   \\ \hline
   B &

  \begin{lstlisting}
   {
	"instruction": 
	'''
	Based on the understanding of the input content and the candidate answers provided in the choice, answer the question. Note that the generated answer must come from the choices, directly output the answer without any extra content. 

    input: 
        Woman: What are you writing? A diary?
        Man: No, it's a semester plan. I write this kind of plan before the start of every semester. It's become a habit for me.
    question: What is the man writing?
    choice: ["diary", "novel", "semester plan", "work summary"]
	''',
	"output": "semester plan"
}

  \end{lstlisting}
  \\ \hline
  C &

  \begin{lstlisting}
  {
	"instruction": 
    '''
    	Based on the understanding of the input and the candidate answers provided in choice, answer the question. Note that the generated answer must come from the choices, directly output the answer without any extra content.
    examples: 
        input: 
            Female: Why don't you pay attention in class? Is there something wrong?
            Male: I just can't focus! I'm not cut out for studying, but my parents just don't listen.
        question: According to the conversation, what is the male like?
        choice: ["Can't understand the teacher's lectures", "Doesn't want to listen to parents", "Lacks interest in studying", "Thinks the content is too simple"]
        answer: Lacks interest in studying
    
        input: In our country, the second Friday of August every year is "Take Your Child to Work Day." On this day, children can come to work with us and understand how hard our work is.
        question: Why do we take our children to work?
        choice: ["For fun", "To understand our work", "No one to look after the children"]
        answer: To understand our work
    
    input: 
        Woman: What are you writing? A diary?
        Man: No, it's a semester plan. I write this kind of plan before the start of every semester. It's become a habit for me.
    question: What is the man writing?
    choice: ["diary", "novel", "semester plan", "work summary"]
	''',
	"output": "semester plan"
}
  \end{lstlisting}
 \\ \hline
 \end{tabular}
  \caption{Instruction Example of C3} 
  \label{Tab.C3}
  \end{table}

   \begin{table}
   \begin{tabular}[c]{p{1 cm}|p{15cm}}
   \toprule
    Task & Instruction 
   \\ \hline
   B &
  \begin{lstlisting}
{
	"instruction": "Please analysis the emotional tendency reflected in the review text in the input,  directly output \"Positive\" or  \"Negative\" without any additional content.
    input:
        wow, this movie sucked.<br /><br />This movie was a embarrassment to the original sandlot.<br /><br />Everything about this movie was awful.<br /><br />The acting was horrendous. Every part except the part of the 'mexican' sandlot manager was terrible.<br /><br />Luke Perry, though only bit parts was absolutely awful. This was is worst role ever. Even the kid actor playing him as a kid was someone you'd want to punch, even in the end, lol.<br /><br />This movie reminded me of those kid movies that go that extra mile making a part goofy way beyond the funny stage. The humor was for 6 year olds.<br /><br />If your over 12 and want something worthwhile to watch, skip this movie and watch a sitcom instead.",
	"output": "Negative"
}
  \end{lstlisting}
  \\ \hline
  C &
  \begin{lstlisting}
{
	"instruction": "Please analysis the emotional tendency reflected in the review text in the input, then choose Positive or Negative within their description.  Please directly output \"Positive\" or  \"Negative\" without any additional content.
    Positive: The overall tone of the review is positive, indicating satisfaction or enjoyment. Language Used: Words and phrases are favorable, enthusiastic, or appreciative. Uses terms like amazing, wonderful, incredible, excellent, loved it, etc.
    Negative: The overall tone of the review is negative, indicating dissatisfaction or disappointment. Language Used: Words and phrases are critical, disapproving, or unimpressed.Uses terms like terrible, awful, disappointing, bad, hated it, etc.
    input: wow, this movie sucked.<br /><br />This movie was a embarrassment to the original sandlot.<br /><br />Everything about this movie was awful.<br /><br />The acting was horrendous. Every part except the part of the 'mexican' sandlot manager was terrible.<br /><br />Luke Perry, though only bit parts was absolutely awful. This was is worst role ever. Even the kid actor playing him as a kid was someone you'd want to punch, even in the end, lol.<br /><br />This movie reminded me of those kid movies that go that extra mile making a part goofy way beyond the funny stage. The humor was for 6 year olds.<br /><br />If your over 12 and want something worthwhile to watch, skip this movie and watch a sitcom instead.\nanswer:",
	"output": "Negative"
}
  \end{lstlisting}
 \\ \hline
 \end{tabular}
  \caption{Instruction Example of IMDB} 
  \label{Tab.C3}
  \end{table}


   \begin{table}
   \begin{tabular}[c]{p{16 cm}}
   \toprule
    an instruction of OpenIE task
   \\ \hline
  \begin{lstlisting} 
{
	"instruction": "You are an expert in open information extraction. Below is a text. Please extract the elements of subject, predicate, object, time, and location from the text. Return them in the format: {\"subject\":[subject], \"predicate\":[predicate], \"object\":[object], \"time\":[time], \"location\":[location]}, arranged in the order they appear in the text. Do not output elements that do not exist.\nInput:Defoe 's A Review , published on 3 December 1709 and demanding `` a Law in the present Parliament ... for the Encouragement of Learning , Arts , and Industry , by securing the Property of Books to the Authors or Editors of them '' , was followed by How 's Some Thoughts on the Present State of Printing and Bookselling , which hoped that Parliament `` might think fit to secure Property in Books by a Law '' .",
	"output": "(\"Defoe\":[subject], \"'s\":[predicate], \"A Review , published on 3 December 1709 and demanding `` a Law in the present Parliament ... for the Encouragement of Learning , Arts , and Industry\":[object])\n(\"A Review\":[subject], \"published\":[predicate], \"on 3 December 1709\":[object])\n(\"Some Thoughts on the Present State of Printing and Bookselling\":[subject], \"hoped\":[predicate], \"that Parliament `` might think fit to secure Property in Books by a Law\":[object])\n(\"Parliament\":[subject], \"might think\":[predicate], \"fit to secure Property in Books by a Law\":[object])"
}
   
  \end{lstlisting}
    \\ \hline
 \end{tabular}
  \caption{Instruction Example of OpenIE} 
  \label{Tab.OpenIE}
\end{table}

\begin{table}
   \begin{tabular}[c]{p{16 cm}}
   \toprule
    an instruction of TC task
   \\ \hline
  \begin{lstlisting}
 {
	"instruction": "Please classify the topic of the text in input and choose the type within the scope defined in the schema.",
	"schema": ["Constellation, entertainment, technology, society, stocks, real estate, education, lottery, home decoration, games, current affairs, fashion, sports"],
	"input": "Bright single: Member 48 yuan wins the first prize in the double color ball, the first cold is fully covered (picture)\n Beijing time, May 3, 2010, the 10044th issue of the double color ball lottery was announced. The lottery result was relatively positive. The first prize had 1033 winners, each winning 13278 yuan, the second prize had 329 yuan, and the first prize for selecting any nine games was 157 yuan. \n\n",
	"output": "{\"type ": "lottery\"}"
}
   
  \end{lstlisting}
    \\ \hline
 \end{tabular}
  \caption{Instruction Example of TC} 
  \label{Tab.TC}
\end{table}

   \begin{table}
   \begin{tabular}[c]{p{16 cm}}
   \toprule
    an instruction of KGE task 
   \\ \hline
  \begin{lstlisting}%[language=json,firstnumber=1]

{
	"task": "KGE",
	"instruction": {
		"instruction": "You are an expert in structured knowledge systems for graph entities. Based on the schema description of the input entity type, you extract the corresponding entity instances and their attribute information from the text. Attributes that do not exist should not be output. If an attribute has multiple values, a list should be returned. The results should be output in a parsable JSON format.",
		"schema": [{
			"entity_type": "Works",
			"attributes": ["achievement", "director", "performer", "lyrics by", "composer", "platform", "screenwriter", "author", "developer", "based on", "country of origin", "tracklist", "publisher", "production company", "box office", "original broadcaster", "cast member"]
		}],
		"input": "The Lego Batman Movie  is the soundtrack to the 2017 computer-animated film The Lego Batman Movie, which is the second instalment in The Lego Movie franchise. The film is based on the DC Comics superhero Batman, and other primary characters from the DC Universe and the Lego DC Super Heroes' Batman toy line. This is the first and only film in the franchise not to be scored by Mark Mothersbaugh, instead Lorne Balfe scored for the film.  The soundtrack to the film was released by WaterTower Music, through two-disc CD formats and for digital download, on February 3, 2017, a week prior to the film's release. A vinyl edition of the soundtrack was released on May 19, 2017."
	},
	"output": {
		"Works": {
			"The Lego Batman Movie": {
				"composer": "Lorne Balfe"
			}
		}
	}
}

  
\end{lstlisting}
  \\ \hline
 \end{tabular}
  \caption{Instruction Example of KGE} 
  \label{Tab.KGE}
\end{table}

\begin{table}
   \begin{tabular}[c]{p{16 cm}}
   \toprule
    an instruction of NER task
   \\ \hline
  \begin{lstlisting}%[language=json,firstnumber=1]

{
	"instruction": {
		"instruction": "You are an expert in named entity recognition. Please extract entities that match the schema definition from the input. Return an empty list if the entity type does not exist. Please respond in the format of a JSON string.",
		"schema": ["average ratings", "year", "title", "actor", "character", "song"],
		"input": "please show me a documentary featuring jessica lange from the 2010 s"
	},
	"output": {
		"average ratings": [],
		"year": ["2010 s"],
		"title": [],
		"actor": ["jessica lange"],
		"character": [],
		"song": []
	}
}

\end{lstlisting}
   \\ \hline
 \end{tabular}
  \caption{Instruction Example of NER.} 
  \label{Tab.NER}
  \end{table}

   \begin{table}
   \begin{tabular}[c]{p{16 cm}}
   \toprule
    
    an instruction of RE task 
   \\ \hline
  \begin{lstlisting}%[language=json,firstnumber=1]

{
	"instruction": {
		"instruction": "Please extract the elements that match the schema definition from the input and return the results in the format specified in the output_format.",
		"schema": ["country of capital", "children", "country of administrative divisions", "ethnicity"],
		"output_format": {"predicate": [{"subject": "", "object": ""}]},
		"input": "At a meeting in Montevideo , Uruguay , the four members of the trade bloc -- Brazil , Argentina , Paraguay and Uruguay -- are expected to formally begin negotiations to bring Venezuela into Mercosur , a group that seeks to standardize tariffs and trade practices throughout the region ."
	},
	"output": {
		"country of capital": [{"subject": "Uruguay", "object": "Montevideo"}],
		"children": [],
		"country of administrative divisions": [],
		"ethnicity": []
	}
}

    \end{lstlisting}
       \\ \hline 
    \end{tabular}
  \caption{Instruction Example of RE} 
  \label{Tab.RE}
\end{table}

   \begin{table}
   \begin{tabular}[c]{p{16 cm}}
    \toprule
       Instruction of SPO task
  \\ \hline  
  \begin{lstlisting}%[language=json,firstnumber=1]

{
	"instruction": {
		"instruction": "You are an expert specializing in the extraction of SPO triplets. Please extract triplets from the input that conform to the defined schema. Return an empty list for relationships that do not exist. Please respond in the format of a JSON string. You can refer to the example for extraction.",
		"schema": [{
			"subject_type": "disease",
			"predicate": "related (caused by)",
			"object_type": "disease"
		}],
		"input": "The characteristics of schistosomiasis include symptoms of the hepatobiliary system (such as abdominal pain, jaundice, right upper abdominal pain), pulmonary symptoms (such as chronic cough, chest pain, dyspnea and hemoptysis) or digestive symptoms (such as mucosal ulcers, malnutrition)."
	},
	"output": {
		"related (caused by)": [
            {"subject": "schistosomiasis","object": "jaundice"},
			{"subject": "schistosomiasis","object": "mucosal ulcers"},
			{"subject": "schistosomiasis","object": "malnutrition"}
		]
	}
}
    \end{lstlisting}

  \\ \hline 
    
 \end{tabular}
  \caption{Instruction Example of SPO.} 
  \label{Tab.SPO}
    \end{table}

  \begin{table}
   \begin{tabular}[c]{p{16 cm}} 
      \toprule
      Instruction of MRC task
    \\ \hline 
  \begin{lstlisting}
{
	"instruction": {
		"instruction": "Please answer the question in question based on the content in input. If there is no answer in input, return: Not mentioned.",
		"input": "2. Megatron: The cold leader of the Decepticons, the main antagonist in 'Transformers'.",
		"question": "What is the name of the antagonist in 'Transformers'?"
	},
	"output": {
		"answer": "Megatron"
	}
}
  \end{lstlisting}
 \\ \hline
 \end{tabular}
  \caption{Instruction Example of MRC} 
  \label{Tab.MRC}
  \end{table}

\begin{table}
   \begin{tabular}[c]{p{16. cm}}
    \toprule
    Instruction of EE task
   \\ \hline
   \begin{lstlisting}%[language=json,firstnumber=1]

{
	"instruction": {
		"instruction": "You are an expert in event extraction. Please extract events from the input that conform to the schema definition. Return an empty list for events that do not exist, and return NAN for arguments that do not exist. If an argument has multiple values, please return a list. Respond in the format of a JSON string.",
		"schema": [{
			"event_type": "data breach",
			"trigger": true,
			"arguments": ["number of victim", "number of data", "purpose", "attacker", "compromised data", "victim", "place", "time", "attack pattern", "tool", "damage amount"]
		}, {
			"event_type": "ransom",
			"trigger": true,
			"arguments": ["damage amount", "place", "victim", "payment method", "attack pattern", "attacker", "time"]
		}],
		"input": "Leading French presidential candidate Emmanuel Macron's campaign said on Friday it had been the target of a `` massive'' computer hack that dumped its campaign emails online 1-1/2 days before voters choose between the centrist and his far - right rival , Marine Le Pen ."
	},
	"output": {
		"data breach": [{
			"trigger": "hack",
			"arguments": {
				"number of victim": "NAN",
				"number of data": "NAN",
				"purpose": "NAN",
				"attacker": "NAN",
				"compromised data": "NAN",
				"victim": "computer",
				"place": "NAN",
				"time": "Friday",
				"attack pattern": "NAN",
				"tool": "NAN",
				"damage amount": "NAN"
			}
		}],
		"ransom": []
	}
}

\end{lstlisting}

\\ \hline 

\end{tabular}
  \caption{Instruction Example of EE.} 
  \label{Tab.EE}
    \end{table}

\begin{table}
   \begin{tabular}[c]{p{16.cm}}
      \toprule
    Instruction of EET task 
   \\ \hline
   \begin{lstlisting}
{
	"instruction": {
		"instruction": "You are an expert in event extraction. Please extract event types and event trigger words from the input that conform to the schema definition. Return an empty list for non-existent events. Please respond in the format of a JSON string.",
		"schema": {
			"nominate": "'Nominate' selects candidates for job or honor; trigger words include 'nominations', 'named', 'selecting', 'nomination'.",
			"attack": "An 'attack' event is an attempt to harm indicated by trigger words in a text, even if not yet carried out.",
			"phone write": "Event emphasizing communication through phone calls, emails, messages. Can be formal or informal. Trigger words: 'Call', 'email', 'message'.",
			"transport": "Moving or transporting something or someone from one place to another. Includes relocating, deploying resources, and lifting off.",
			"label81": "'Convict' means being declared guilty of a crime, leading to penalties. It can happen formally or informally. Trigger words include 'found', 'pled guilty', 'convicted'."
		},
		"input": "a member of the international committee of red cross visited the local hospital there , and he says it ' s a horrible scene ."
	},
	"output": {
		"nominate": [], "attack": [], "phone write": [], "transport": ["visited"], "label81": []
	}
}
\end{lstlisting}

\\ \hline 

\end{tabular}
  \caption{Instruction Example of EET.} 
  \label{Tab.EET}
    \end{table}

\begin{table}
   \begin{tabular}[c]{p{16.cm}}
      \toprule
    Instruction of EEA task
   \\ \hline
   \begin{lstlisting}
{
	"instruction": {
		"instruction": "You are an expert in event argument extraction. Please extract event arguments and their roles from the input that conform to the schema definition, which already includes event trigger words. If an argument does not exist, return NAN or an empty dictionary. Please respond in the format of a JSON string.",
		"schema": [{
			"event_type": "adverse event",
			"arguments": ["Treatment.Dosage", "Subject.Age", "Treatment.Drug", "Treatment.Disorder", "Treatment.Route", "Treatment.Time_elapsed", "Subject.Gender", "Treatment.Freq", "Effect", "Treatment", "Subject.Race", "Combination.Drug", "Subject.Population", "Subject", "Subject.Disorder"]
		}],
		"input": "CONCLUSION: Fixed drug eruption is associated with many drugs but this is the first such report with omeprazole."
	},
	"output": {
		"adverse event": [{
			"Treatment.Dosage": "NAN",
			"Subject.Age": "NAN",
			"Treatment.Drug": "omeprazole",
			"Treatment.Disorder": "NAN",
			"Treatment.Route": "NAN",
			"Treatment.Time_elapsed": "NAN",
			"Subject.Gender": "NAN",
			"Treatment.Freq": "NAN",
			"Effect": "Fixed drug eruption",
			"Treatment": "omeprazole",
			"Subject.Race": "NAN",
			"Combination.Drug": "NAN",
			"Subject.Population": "NAN",
			"Subject": "NAN",
			"Subject.Disorder": "NAN"
		}]
	}
}
\end{lstlisting}
\\ \hline 



 \end{tabular}
   \caption{Instruction Example of EEA.} 
  \label{Tab.EEA}
    \end{table}

\begin{table}
   \begin{tabular}[c]{p{16cm}}
   \toprule
    An EE task  instruction with guidelines, the description of the event type together with the   argument description and typical\_example for each argument are given.
   \\ \hline
\begin{lstlisting}
{
	"instruction": {
		"instruction": "You are an expert in event argument extraction. Please extract event arguments and their roles from the input that conform to the schema definition, which already includes event trigger words. If an argument does not exist, return NAN or an empty dictionary. Please respond in the format of a JSON string.",
		"schema": {
			"discover vulnerability": {
				"discover vulnerability": "Event type for identifying and reporting software or system weaknesses, often triggered by certain words.\ntypical examples: have had their fair share",
				"arguments": [
					{"type": "vulnerable system version","description": "Vulnerable system versions can vary from specific ones like '3.2.2' to ranges like '2.3.5-2.3.31' or versions before '6.5.2'.","typical examples": "versions 2.5"},
					{"type": "time","description": "When was the vulnerability discovered? It could be a date, time, month, or year.Typical examples: last Saturday"},
					{"type": "supported platform","description": "Argument role in 'discover vulnerability' event: technology, software/hardware platforms, operating systems, services, hardware devices susceptible to cyberattacks.Typical examples: both iOS and Android devices"},
					{"type": "vulnerability","description": "System owner identifies responsible entities for systems, software, or hardware with security vulnerabilities, including WordPress, Magento, D-Link, vendors, Technicolor, telecom, tech companies, TVT Digital, and Splunk. Typical examples: Four Cross-Site Scripting (XSS) vulnerabilities"},
					{"type": "capabilities","description": "Discovered vulnerabilities can allow unauthorized access, data leaks, system crashes, and control of devices without passwords. Typical examples: using the PROPFIND method and IF header"},
					{"type": "common vulnerabilities and exposures","description": "CVE standardizes identifying cybersecurity vulnerabilities with unique identifiers like 'CVE-2017-7692'. Widely used in the security industry. Typical examples: CVE -2017-3823"},
					{"type": "vulnerable system","description": "Tech prone to exploitation: systems like Microsoft Word, Windows Server 2003 R2, Firefox, routers, Magento, smart watches, etc. Typical examples: widely used services"}
				]
			}
		},
		"input": "Two days later , Timothy Morgan of Blindspot Security came forward and presented a more ominious exploitation scenario where the FTP URL handlers in Java and Python could be used to bypass firewalls ."
	},
	"output": {
		"discover vulnerability": [{
			"vulnerable system version": "NAN",
			"time": "Two days later",
			"supported platform": "NAN",
			"vulnerability": "a more ominious exploitation scenario",
			"capabilities": "be used to bypass firewalls",
			"common vulnerabilities and exposures": "NAN",
			"vulnerable system": ["Java", "Python", "the FTP URL handlers"]
		}]
	}
}
\end{lstlisting}\\ \hline 
\end{tabular}
  \caption{Instruction Example of Guidelines.} 
  \label{Tab.guideline}
\end{table}

\begin{table}
   \begin{tabular}[c]{p{16cm}}
   \toprule
    Two instructions synthesised from the same sample of NER task with different preference rules.
   \\ \hline
\begin{lstlisting}
{
	"instruction": {
		"instruction": "You are an expert in named entity recognition. Please extract entities that match the schema definition from the input. Return an empty list if the entity type does not exist. Please respond in the format of a JSON string.",
		"schema": [{
			"entity_type": "money",
			"description": "Represents monetary values, such as income, prices, or asset amounts.",
			"rule": "Extract all monetary values mentioned in the text, If there are units ot symbols before or after the numerical value, please ignore them."
		}],
		"input": "Average household income for the sample was $ 194,000 , and average net assets were reported as $ 775,000 ."
	},
	"label": [
        {"entity_type": "money", "entity": "194,000"},
        {"entity_type": "money","entity": "775,000"}
    ]
}
\end{lstlisting}\\ \hline 
\begin{lstlisting}
{
	"instruction": {
		"instruction": "You are an expert in named entity recognition. Please extract entities that match the schema definition from the input. Return an empty list if the entity type does not exist. Please respond in the format of a JSON string.",
		"schema": [{
			"entity_type": "money",
			"description": "Represents monetary values, such as income, prices, or asset amounts.",
			"rule": "Extract all monetary values mentioned in the text, If there are units ot symbols before or after the numerical value, please also extract them together."
		}],
		"input": "Average household income for the sample was $ 194,000 , and average net assets were reported as $ 775,000 ."
	},
	"label": [
        {"entity_type": "money", "entity": "$ 194,000"},
        {"entity_type": "money","entity": "$ 775,000"}
    ]
}


\end{lstlisting}\\ \hline 
\end{tabular}
 \caption{Instruction Example of Preference Rule.} 
 \label{Tab.Rule}
\end{table}

\begin{table}
   \begin{tabular}[c]{p{16cm}}
   \toprule
     Two instructions synthesised from the same sample of RE task with format variants.
   \\ \hline
\begin{lstlisting}
{
	"instruction": "Please extract the elements that match the schema definition from the input and return the results in the format specified in the output_format ",
	"output_format": {
		"predicate": [{"subject": "", "object": ""
		}]
	},
	"schema": ["Kill"],
	"output": {
		"kill": [{"subject": "Sirhan Sirhan", "object": "Robert F. Kennedy"}]
	}
}
\end{lstlisting}

\\ \hline
\begin{lstlisting}
 
{
	"instruction": "Please extract the elements that match the schema definition from the input and return the results in the format of markdown Table.The header is | subject | predicate | object |",
	"schema": ["Kill"]
	"output": "| subject |predicate | object |\n| --- | --- |--- |\n| Sirhan Sirhan| kill | Robert F. Kennedy |"
}

\end{lstlisting}
\\ \hline 



 \end{tabular}
  \caption{Instruction Example of format variants.} 
  \label{Tab.format}
\end{table}

\end{document}